\documentclass{article}
\usepackage{iclr2027_conference,times}

\usepackage{amsmath,amsfonts,bm}

\def\eqref#1{equation~\ref{#1}}

\def\1{\bm{1}}

\DeclareMathAlphabet{\mathsfit}{\encodingdefault}{\sfdefault}{m}{sl}
\SetMathAlphabet{\mathsfit}{bold}{\encodingdefault}{\sfdefault}{bx}{n}

\usepackage[hidelinks]{hyperref}
\usepackage{url}
\usepackage{booktabs}
\usepackage{graphicx}
\usepackage{amsmath,amssymb}
\usepackage{caption}
\usepackage{xspace}
\providecommand{\eg}{e.g.}
\providecommand{\ie}{i.e.}
\newcommand{\spark}{SPARK\xspace}
\newcommand{\lib}{LIBERO\xspace}
\newcommand{\libpro}{LIBERO-PRO\xspace}
\newcommand{\capx}{CaP-Agent0\xspace}
\newcommand{\capbench}{CaP-Bench\xspace}
\newcommand{\rats}{RATS\xspace}

\title{Sequential Planning via Anchored Robotic Keypoints}

\author{
  Bryce Grant$^{1}$ \quad
  Aryeh Rothenberg$^{2}$ \quad
  Logan Senning$^{2}$ \\
  \bfseries Zonghe Chua$^{1}$ \quad
  Zach Patterson$^{2}$ \quad
  Peng Wang$^{2}$ \\
  \\
  $^{1}$ Department of Electrical, Computer, and Systems Engineering \\
  $^{2}$ Department of Mechanical and Aerospace Engineering \\
  Case Western Reserve University, United States \\
  \texttt{bag100@case.edu}, \texttt{pxw206@case.edu}
}

\iclrfinalcopy
\begin{document}
\maketitle
\begin{abstract}
Changes to what a robot can see do not mean that there is a change in the problem it is trying to solve, and many language-conditioned manipulation systems struggle with this distinction. When something in a scene moves, the robot might need to execute its actions in different places, but its manipulation plan can stay the same. We propose Sequential Planning via Anchored Robotic Keypoints (\spark), which describes a task as a symbolic behavior tree that we anchor to the current 3D observations when each action is called. By verifying the plan at runtime we can recover from failures locally and anchor the objects to the scene again. We ask for a new plan only if this fails. We isolate this concept by taking a fixed plan and moving objects in the scene while executing. When displacing objects by $10$\,cm mid-episode we were still able to complete $60.0\%$ of trials with scene-diff verification and layered recovery, and we did not have to change the plan. Without practicing on the tasks being tested or using ground-truth object poses from the simulator during rollouts, we test our system (with one open-weight 27B planner with 4-bit weights) on \libpro. Under this protocol \spark achieves a mean success of $74.4\%$ across the 6 position and task settings when the robot decides to retry from its own checks, and $77.8\%$ when it also reads a success check after each attempt, as in prior work. We also achieve a $75\%$ success rate across 11 physical task-embodiment settings on 3 different robots. Our work demonstrates that for an important class of failures, the robot does not need to ask for a new plan for its task. It can be enough to update the system's view of the world. Code is at \url{https://github.com/CWRU-AISM/sparkbot} and videos are at \url{https://cwru-aism.github.io/spark-page/}

\end{abstract}
\section{Introduction}
\label{sec:intro}

General-purpose robot policies will have to be able to account for variations in their environment and in what they are trying to do. There are variations that will affect the task a robot is trying to accomplish and there are variations that only affect where that task takes place. If a bowl is moved on a table, the robot's sensing and the actions needed to place the bowl on the plate may change, but the task itself often does not. We want our robot to maintain the aspects of the task that can stay the same. When operating in the real world we will constantly see variation in how objects are placed in the scene, so our manipulation system needs to account for this.

Vision-language-action models (VLAs) currently don't explicitly separate the two. OpenVLA, \(\pi_0\) and \(\pi_{0.5}\) are end-to-end models that predict robot actions from camera images and a language instruction, and \(\pi_0\) and \(\pi_{0.5}\) also take the robot state as input~\citep{kim2024openvla,black2024pi0,physicalintelligence2025pi05}. Evaluations of the robustness of these models show that their performance depends strongly on the scenes and tasks seen in training. The \libpro benchmark~\citep{zhou2025liberopro} showed that these models score above 90\% on the original LIBERO benchmark~\citep{liu2023libero} but nearly collapse when object positions or tasks are changed. Another work~\citep{grant2026sae} looked at the internal representations of several VLAs and found that their motor representations become tied to locations in a scene. If you take those internal activations from one task and apply them to another task you can steer the model to follow the path and location of the source task. When something changes about the state of the world it can break the actions a policy takes, even when the plan for the task is still appropriate.

There have been works that try to add reasoning into the loop at test time. Works like \capx generate code that the robot can execute. They use multiple interactions with an LLM, execution feedback and differences in images after executing actions to correct failures~\citep{fu2026capx}. More recent works have achieved high performance on the \libpro benchmark by allowing agents to practice on the evaluated tasks before they are scored, or by thinking for many minutes per episode while the simulator is paused (\S\ref{sec:related}). While these are promising approaches they leave us with a question: when we fail to complete a task, what do we need to recompute?

We believe that in order to have a robust manipulation system we need to know what information should update and what should not. If the instructions change the robot may need to rethink its plan. If an object moves the robot may only need to update where it thinks that object is located. There could also be failures due to the limits of the robot's execution skills. We should not have to rethink our entire plan when our robot is unable to grasp an object correctly. So if our task structure can remain the same we should be able to perceive and execute actions on that scene without needing to solve the task again.

In this work we propose Sequential Planning via Anchored Robotic Keypoints (\textbf{SPARK}). Our method separates the plan from where objects are located in the scene. We use an LLM to create a symbolic behavior tree for the task. Our behavior tree consists of a fixed set of skills that our robot can execute. Once we have our plan we know what objects we want to interact with and what actions we want to take, but not where those objects are. Then when it comes time to execute an action we can detect that object and anchor it to its current 3D location in our scene. We check if the action we wanted to take was successful. If it was not, we can go back and detect that object again and try the action once more with the new location, while the rest of the plan stays the same. We only need to create a new plan if we cannot recover from our mistake. This allows our task to stay the same while we update our anchors in the scene.

We evaluate this separation through controlled object-displacement experiments, LIBERO-PRO benchmarks, physical robot experiments, and ablations of planning and perception. Our contributions include:
\begin{enumerate}
    \item \textbf{Separating symbolic plans from execution-time grounding.} Our proposed approach, SPARK, represents a manipulation task as a symbolic behavior-tree while grounding its object references to current 3D observations only when the action is executed. Using runtime post-condition checks, this allows robust execution under changes in object position without requiring the symbolic plan to be regenerated. Every executed step is a typed node with a checked outcome, so each rollout is also recorded as labelled segments (\S\ref{sec:discussion}).
    \item \textbf{Controlled evidence for recovery without replanning.} We introduce a scripted displacement protocol to test whether execution can recover when objects move during a task (LIBERO-Displace, \S\ref{sec:libero_dyn_main}). With the plan pinned in place, we find that scene-diff verification and layered recovery allow us to succeed in $60.0\%$ of trials at 10\,cm displacement. These results identify failure modes where execution can be recovered by re-grounding the plan's primitives while maintaining the same underlying plan.
    \item \textbf{\libpro and physical results.} With no practice on the evaluated tasks, no ground-truth object poses read from the simulator, and one open-weight 27B planner with 4-bit weights, SPARK reaches $77.8\%$ mean success on six \libpro position and task settings when a success check after each attempt decides retries, as in \rats, and $74.4\%$ when only its own checks decide. On the held-out language and object settings it reaches $69.7\%$. SPARK succeeds in $75\%$ of trials across eleven physical task-embodiment settings on three robot platforms (\S\ref{sec:libero_main}, \S\ref{sec:realworld}).
    \item \textbf{Characterizing when additional perception improves execution.} Alternative object descriptions improve success on the LIBERO-PRO Spatial and Object suites by $26.5$ and $8.2$ percentage points, respectively, while Goal-task drops by $8.4$ points (\S\ref{sec:matched_budget}).
\end{enumerate}

\section{Problem Setup}
\label{sec:problem}

We study language-conditioned manipulation when object placements and task instructions change. Given an instruction and observations from a camera we want our robot to perform a set of actions that complete the desired task. We utilize existing models and a fixed skill library to complete tasks. No models are trained for a specific task.

We define an \emph{anchored robotic keypoint} as a 3D point that has been grounded to an object/part label using open vocabulary detection. SAM3 is used to find the named object or part (e.g., a bowl or drawer handle) and the detection is then back projected into world space using depth. The label is used to define what we are looking for and the keypoint location can be changed with new detections. This allows our executor to re-use the program when an object is moved but the actions are still valid. If there is a change to the instruction the executor may need to target different objects or take different actions. A keypoint is just where we want to target. Skills and controllers will decide how we get to that point and how we interact with it.

We test changing object placements and instructions during different phases of an episode and also have scripted movements of objects while executing. Our robot will react to these changes using observations from the current episode without updating model weights.

There can be mistakes made by choosing the wrong actions or by grounding objects incorrectly. We can also run into issues where our execution skill cannot perform a certain task. Mistakes in planning will result in choosing bad actions or targets. Grounding mistakes will happen when an object label is applied to the wrong object or location. All of these issues can overlap with each other. Our post condition will tell us what step failed but not, by itself, what caused it. We explore if more perception helps our success rate in these scenarios (\S\ref{sec:matched_budget}).

We define evaluation on a task with no knowledge of past attempts and with knowledge of past attempts. During evaluation on a task with no knowledge of past attempts our robot is not given any examples or memories from a task. Evaluation with knowledge of past attempts can allow our robot to perform better on future trials by having saved programs/examples/memories about a task. The main \spark evaluation is done without retrieval and write back of successful behavior trees. We also distinguish whether simulator state or success signals are available within the experimental setup.

Our research question is: to what extent can we leverage programs, runtime verification and current foundation models to perform tasks, without any bespoke learning per task? Although \spark does not use a VLA for task execution, it does assume pretrained models and a skill library, as well as control calibration for each platform.

\section{Related Work}
\label{sec:related}

\paragraph{LLM-to-skill planning and BT synthesis.}
Given a flat menu of pretrained skills, SayCan~\citep{ahn2022saycan} generates a sequence of them. In contrast, we use a typed grammar as the interface and emit one behavior tree per task that is directly executable, with no learned policy at runtime. Related approaches synthesize BTs with LLMs: \citet{chen2024llmobtea} convert a natural-language instruction to a first-order logic goal and build an optimal BT for it with a planning algorithm and \citet{salfity2026contractbt} validate generated BTs against a robot-side contract of available skills and typed parameters before execution. We go further and check post-conditions of primitives at execution, leveraging perception to re-ground the objects on a failed check, again without querying the LLM.

\paragraph{VLA fragility and learned executors.}
OpenVLA~\citep{kim2024openvla}, $\pi_0$~\citep{black2024pi0} and $\pi_{0.5}$~\citep{physicalintelligence2025pi05} achieve $>$90\% success on standard LIBERO, but performance drops precipitously when positions or tasks are changed in \libpro~\citep{zhou2025liberopro} (\S\ref{sec:intro}). MolmoAct2~\citep{fang2026molmoact2} scores 97.2\% on standard LIBERO, but in our evaluation of its released checkpoint it only achieves 18.6\% on average across the six perturbed \libpro cells (Table~\ref{tab:libero_main}), suggesting that it has memorized poses.

\paragraph{Code-generation agents and recovery.}
Code-as-Policies~\citep{liang2023codepolicies} generates Python code using robot APIs and ProgPrompt~\citep{singh2023progprompt} includes assertions to ensure preconditions are met. In our method we make sure that the post condition of a primitive is true using a re-detected scene during runtime. \citet{colledanchise2018bt} describe postcondition, precondition, action structures in behavior trees. Code-as-Monitor~\citep{zhou2025codeasmonitor} demonstrates a VLM can write monitor code to verify spatio-temporal constraints over tracked geometric primitives both during and after execution to detect failures. We believe our use of types to link a postcondition to a primitive gives us the ability to pinpoint where a failure occurred. \capx~\citep{fu2026capx} builds on this line with multi-turn code generation, an ensemble of 3 different models and visual differencing. ReKep~\citep{huang2024rekep} also uses keypoints and needs no task-specific model training, like \spark, and it re-solves a sequence of end-effector poses at ${\sim}10$\,Hz.

\paragraph{Failure detection and recovery.}
PLANEX re-executes the failed part of a plan before it replans~\citep{fikes1972planex}, and RAP leaves steps to be chosen as the situation unfolds~\citep{firby1989rap}. Perceptual anchoring creates and maintains the link between a symbol and the sensor data of the same object~\citep{coradeschi2003anchoring}, and our keypoints follow this idea. Sentinel detects failures of generative policies from the temporal consistency of their actions and from VLM progress checks~\citep{agia2024sentinel}. \citet{sinha2024aesop} pair a fast anomaly classifier in an LLM embedding space with slower LLM reasoning that selects a safe fallback. Fail2Progress generates simulation data close to observed failures and fine-tunes a skill-effect model on it~\citep{huang2025fail2progress}. REFLECT~\citep{liu2023reflect}, DoReMi~\citep{guo2024doremi} and Recover~\citep{cornelio2024recover} detect a failure and use a language model to explain it or to replan. \citet{vlmbtfailure2025} combine VLMs, a reactive planner and behavior trees to handle failures during execution. Cybo-Waiter has a VLM write typed subtasks with success conditions and checks them with SAM3 and RGB-D geometry~\citep{ren2026cybowaiter}. In our method we keep the plan and detect the objects again, and we measure when that is enough (\S\ref{sec:libero_dyn_main}).

\paragraph{Runtime verification and recovery.}
U-LAG~\citep{jh2025ulag} re-aims task goals during execution with geometric retargeters and leaves the low-level controller unchanged, while \spark checks the post-conditions that its plan and skills define and recovers locally. Harness VLA~\citep{zhang2026harness} has an agentic planner handle semantic grounding and actions where the robot is not in contact with an object, and a frozen VLA handle contact-rich manipulation. They report 82.4\% on \libpro using a Claude Code planner and 72.1\% using Codex. These numbers are achieved with few shot learning, where the agent learns how to complete a task from a reference seed and stores the successful primitive trace. Then during evaluation the agent knows which primitives were successful and re-grounds them. This is evaluation with knowledge of past attempts, as defined in \S\ref{sec:problem}. Harness VLA without any prior task knowledge attains 79.0 on changing goals, dropping to 31.0 when positions are swapped. We attribute its goal score to capabilities that the frozen contact-rich policy provides and analytic primitives lack. The policy they use to interact with objects is an SFT version of $\pi_{0.5}$ trained on robot data, so the comparison is between training-free and pretrained execution.

\paragraph{Practice, simulator access, and cost on \libpro.}
A few recent papers outperform \spark on \libpro with various evaluation setups. On LIBERO-Pro Object, Goal and Spatial, ASPIRE~\citep{lu2026aspire} learns on seeds 51--65 of each task it evaluates on. They score 1 program per task on seeds 1--50. With Claude Opus 5.5 they achieve 98.9\% on the 6 object, goal and spatial settings. It costs them \$10.8 and 46 minutes of agent time per task. EmbodiedRSI~\citep{song2026embodiedrsi} improves code and skills around a frozen VLA model on 10 seeds per task and is tested on another 10 disjoint seeds achieving 86.8\% on 8 settings. Zetta~\citep{zetta2026} improves runtime critics and recovery skills on dev seeds of the tasks being tested. Its diagnosis agent checks object poses and contacts against simulator ground truth, and at runtime its critic reads the official task and grasp predicates. RPG~\citep{wang2026rpg} practices for 15 rounds with a privileged agent that has access to sim state and achieves 95.0\% on unseen initial states of its 22 practice tasks which are not in \libpro. RoboAuto~\citep{liu2026roboauto} gets 89.1\% on the 6 settings by searching for a program for each task over 15 training seeds using Claude Opus~4.6. It sees whether each run succeeded while searching, and then the final program is run on seeds 1--50. These methods all evaluate with knowledge of past attempts (\S\ref{sec:problem}). Agents that operate in the zero-shot category spend their compute at test time. OpenRUA~\citep{chu2026openrua} achieves 91.8\% on the 6 settings in Table~\ref{tab:libero_main} with up to 500 agent turns or 240 minutes per episode. ORA~\citep{yang2026ora} achieves 81.4\% on 8 settings with an open weight model and takes $14$--$25$ minutes per episode on median. Both methods allow the model to think while the sim is stopped. \spark doesn't practice on or search over the tasks it's tested on nor does it read ground-truth object poses from the sim. With one simulation worker it takes a median of $17.9$\,s per episode when only its own checks decide retries (Appendix~\S\ref{sec:appendix:provenance}).

\paragraph{Perturbation benchmarks and test-time compute.}
\libpro~\citep{zhou2025liberopro} perturbs the objects, initial positions, instructions and environment of a task, and applies each perturbation before the episode starts. However, this type of perturbation does not separate methods well, so we also script displacement in the middle of an episode. Three additional LIBERO extensions perturb the scene at runtime as well. LIBERO-MAX~\citep{zhang2026liberomax} executes each scenario twice, once with and once without a single mid-task edit to the scene (e.g. move the target or move the camera), reporting that these edits reduce the performance of fourteen policies by 11.0 to 25.7 points. LIBERO-Dynamic~\citep{ren2026liberodynamic} samples from three families of trajectories while moving objects at three different speeds, while the world continues running forward during model inference time. LIBERO-Pro2.0~\citep{zhou2026liberopro2} pairs twenty runtime edits with twenty-two static scene changes, and reports that policies see an average drop of 26.5 points when faced with runtime edits. CoRe~\citep{yin2026core} applies scripted perturbations mid-episode at phase-indexed stages, and recovers a frozen policy in the test-time setting with no failure examples required. Ours also contrasts, along that axis, a typed post-condition with a learned means of recovery. Further sampling is used in the form of actions (RoboMonkey~\citep{kwok2025robomonkey}) and plans (\capx). In contrast, we use our extra inference budget on perception to re-bind our labels to objects as they move, keeping the original plan intact (\S\ref{sec:matched_budget}). At the same time \rats~\citep{rats2026playful} also spends compute before it sees a task: about $30$M tokens of self-play become a code skill library that it retrieves at test time. However it still writes a program for every trial at test time. \spark keeps no library at test time, so it makes the same number of language-model calls on its 100th task as on its first.

\section{Method}
\label{sec:method}

\begin{figure*}[t]
\centering
\includegraphics[width=0.95\textwidth]{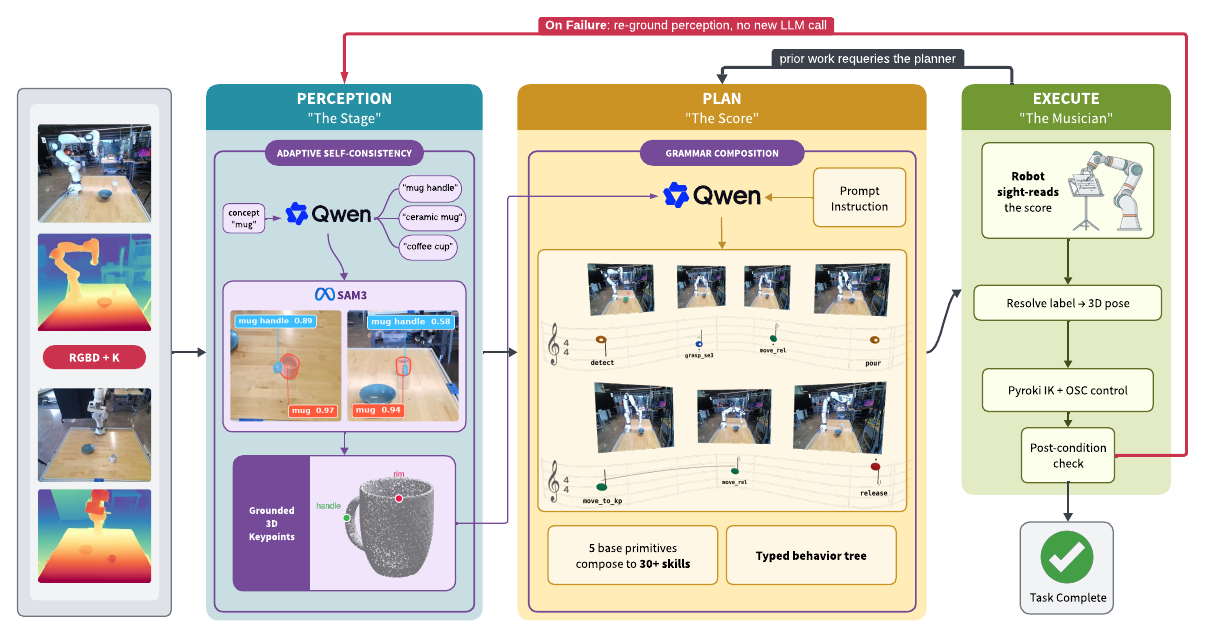}
\caption{\spark decouples a task's score from its execution. Left: perception returns 3D keypoints of the objects. Center: one language-model call writes the score, a behavior tree over a fixed skill library. Right: the robot executes the tree at the current object locations and checks post-conditions. A failed check follows the red path (local correction or re-detection), and a new plan is requested only once the attempt is exhausted.}
\label{fig:architecture}
\end{figure*}

\spark consists of four modules. First we have SAM3 that grounds our scene using the platform cameras. A language model generates a typed behavior tree (BT) over a library of skills that can be combined. We then use an inverse-kinematics (IK) robot controller to perform those actions. If our execution checks fail we can recover with a tiered approach (Figure~\ref{fig:architecture}). Due to its similarity to a musical score, we denote the behavior tree as a score. Our primitives are the notes, their parameters say how to play those notes, and arm assignments coordinate the parts when performing a task with a single arm or both arms. We follow this score and use perception to adapt how we execute it. Our actions point to objects by their label, which we look up at execution time to get the current 3D location. This allows us to try an action again with updated detections without rewriting the score.

\paragraph{Multi-camera perception.}
\label{sec:perception}
Our single arm robots (UR10e, FR3) use a fixed birds eye view camera and a fixed side camera (Azure Kinect, $1920\times 1080$ color, $512\times 512$ time-of-flight depth), along with a wrist camera (RealSense, $640\times 480$ RGB-D). For our bimanual cell we have 1 rear camera and 1 wrist camera per arm. In sim we use the standard camera rig from \libpro (\texttt{agentview} and \texttt{robot0\_eye\_in\_hand}) and do not change its placement. We use SAM3~\citep{carion2025sam3}, an open-vocabulary segmentation model, to segment images using text prompts. We take the centroid of the mask and the median depth value and backproject it into world coordinates using the camera calibration to build a map of detections and associate each object/part label with a 3D location.

\paragraph{Behavior-tree planning.}
\label{sec:planning}
Given our image with SAM3 masks and labels as well as our detected 3D keypoints and instruction we can ask our LM to generate a BT in yaml format. We do this in one prompt. We use a temperature of $0$ for sim and $0.3$ for our hardware results. For Table~\ref{tab:libero_main} we used an open weight model of Qwen3.8-27B quantized to 4bit and ran it locally. We used Gemini for our perception ablation and on our Franka FR3 and bimanual setups, and the same Qwen planner on our UR10e (see the Appendix for which model was used for each program). Each node will point to a registered skill and pass it parameters. We can then verify if our labels are in our detection map. For sim we skip node types that are not in our list of primitives. There are 5 base primitives we use to move around: \texttt{move\_to\_keypoint}, \texttt{move\_relative}, \texttt{grasp}, \texttt{release} and \texttt{wait}. There are over 30 other skills that we have registered that perform calibrated closed loop routines, such as descending until a force is reached, detecting contact and retrying locally. We allow the language model to pick and sequence these skills and pass their parameters, while we handle the lower level control in our executor. Allowing us to reduce the amount of implementation detail, such as coordinate transforms and low-level control, that needs to be generated by systems like \capx that generate code. All of these skills were written using Claude Code against our typed contract and reviewed by the authors of this paper before entering the registry. The registry is fixed for each evaluation and no skills were added or edited during any reported run. This puts \spark in the category of Harness VLA~\citep{zhang2026harness} and Contract-Grounded BT synthesis~\citep{salfity2026contractbt}, which also do not generate code at test time. ASPIRE~\citep{lu2026aspire} writes and repairs a program for each task on debug seeds before it evaluates the program on held-out seeds (\S\ref{sec:problem}). If a behavior tree succeeds we can store it and give it as context in future trials, without updating model weights. However, this was not done for either experimental runs. 

\paragraph{Tiered recovery.}
\label{sec:fallback}
A skill or a plan node can define a post condition that we verify through perception or execution signals. Once we fail that post condition we can identify which step failed and initiate our 3 tiered recovery process. (1) Correct the error in place which may involve a small motion and re-grasping. (2) Re-perceive. Move the arm back $10$\,cm and call SAM3 again and attempt the same plan with new object positions. (3) Call the LLM to generate a new plan after the execution attempt is exhausted. We will use the first two tiers of recovery before changing our plan, and they do not need another LLM call. However, a failed post condition does not always tell us what went wrong. In Table~\ref{tab:libero_main} the planner can write grasp and placement checks into the plan, and they act through the plan's retry and fallback nodes. A grasp check reads the finger gap, so it passes when the gripper holds the wrong object. After a failed attempt we move the arm back $10$\,cm, call SAM3 and get a new plan, at most 2 times. The displacement experiment in \S\ref{sec:libero_dyn_main} uses all three tiers with planner-written plans and the first two with scripted plans.

\paragraph{Alternative-prompt perception.}
\label{sec:adaptive}
When calling SAM3 with a single prompt we often could not find an object or had an incomplete mask. After logging failures we determined this to be the limiting factor on our success rate. Like self consistency~\citep{wang2023selfconsistency}, which samples several reasoning paths for one question, we sample several prompts for each object for our perceptual grounding. We were able to find the object with one of many prompts most of the time. We will make one additional LLM call for every trial to generate $K{=}3$ different prompts for each object. These prompts will describe the object's color, shape and material. For each object if SAM3 can detect the object's name we will use that as our prompt. Otherwise we will take the prompt with the highest confidence score. This confidence score will be reduced if the prompt describes 2 or more objects. When performing recovery we will use the same prompts and not generate new ones. We have two more grounding methods that are used in our \libpro headline. \emph{Wrist-verify}: if we cannot confidently detect our pick target we use the wrist camera at home to verify. Our planner model will choose which numbered crop to use. \emph{Relation-bind}: if there are multiple instances of the same object we will use a static rule to select an instance based on the relation mentioned in the instruction (between, next to, on, in) and the positions of the objects in 3D space. We do not use an LLM call for this method. Our physical experiments only use one prompt per object so they do not test this prompt-generation step (\S\ref{sec:realworld}).

\section{Experiments}
\label{sec:experiments}

\paragraph{Setup.}
We evaluate \spark on \libpro and \capbench with displacement during execution as well as on 3 physical robots. For our \libpro evaluation in Table~\ref{tab:libero_main} we use the 6 position and task settings that were used by \citet{fu2026capx} from \citet{zhou2025liberopro}. These consist of 3 suites (object, goal, spatial) each with 2 variations of perturbations. There are 10 tasks per setting and we run $50$ episodes for each task. Position perturbations move the starting location of objects. Task perturbations may modify the language describing the task or the target object/relation or what must be done depending on the suite. We run 7 Robosuite tasks from \capbench with $100$ episodes each. \spark is equipped with SAM3 and uses simulator depth when in simulation and robot depth when running on physical robots. Controller and IK are kept constant for each robot. When in simulation we use the S2 perception method from \capx so we do not query the simulator for objects. In simulation \spark also uses the known calibration of both cameras, and the VLA rows of Table~\ref{tab:libero_main} use no depth and no camera calibration (inputs per row are in its caption).
 We vary how much information from the simulator and previously successful programs \spark can use in our experiments. In Table~\ref{tab:libero_main} \spark is not given any object poses from the simulator. We use only camera information to verify that pushes, drawers and knobs occurred. No scored episodes are used to practice. Our executor and grounding rules were developed using the position and task settings (\S\ref{sec:limitations}).
The phrase list we use for detection and how far the wrist camera is allowed to move a pick target were picked on re-sampled placements of these settings, which start from the scored initial states and move objects by $0.8$--$3.4$\,cm. Table~\ref{tab:libero_main} uses the $50$ initial states that come with \libpro. We report Table~\ref{tab:libero_main} under two protocols. Under the \rats protocol the success check after each attempt decides whether to retry, as in \rats. Under the strict protocol only the robot's own checks decide whether to retry, and the success check is read once at the end of the episode. Tables~\ref{tab:dyn_main} and~\ref{tab:dyn_matched} use the strict protocol, and the other simulated experiments use the \rats protocol unless stated (Appendix~\S\ref{sec:appendix:exec}). \capbench and both skill ablations also use the Qwen3.8-27B planner of Table~\ref{tab:libero_main} (\S\ref{sec:planning}). No stored programs or cached BTs are used.

\subsection{Performance Under Position and Instruction Changes}
\label{sec:libero_main}

\begin{table*}[t]
\centering
\caption{\libpro success rates ($\%$) on the six position (Pos) and task (Task) settings (\S\ref{sec:libero_main}). OpenVLA, $\pi_0$, $\pi_{0.5}$ and \capx are from \citet{fu2026capx}. \capx (\rats eval) is the run of \capx with only its primitive library reported by \citet{rats2026playful}, alongside \rats. MolmoAct2 is our evaluation of its released LIBERO checkpoint (Appendix~\S\ref{sec:appendix:provenance}). The VLA rows are checkpoints trained with LIBERO demonstrations that take camera images and the instruction ($\pi_0$ and $\pi_{0.5}$ also the robot state); the \spark rows also take rendered depth and camera calibration and use no demonstrations. 95\% intervals of the \spark means: $[71.8, 83.4]$ (\rats protocol) and $[67.5, 80.9]$ (strict). Bold best, underline second-best.}
\label{tab:libero_main}
\small
\setlength{\tabcolsep}{3.5pt}
\begin{tabular}{lccccccc}
\toprule
& \multicolumn{2}{c}{Object} & \multicolumn{2}{c}{Goal} & \multicolumn{2}{c}{Spatial} & \\
\cmidrule(lr){2-3}\cmidrule(lr){4-5}\cmidrule(lr){6-7}
Method & Pos & Task & Pos & Task & Pos & Task & Mean \\
\midrule
OpenVLA~\citep{kim2024openvla} & $0$ & $0$ & $0$ & $0$ & $0$ & $0$ & $0.0$ \\
$\pi_0$~\citep{black2024pi0} & $0$ & $0$ & $0$ & $0$ & $0$ & $0$ & $0.0$ \\
$\pi_{0.5}$~\citep{physicalintelligence2025pi05} & $17$ & $1$ & $38$ & $0$ & $20$ & $1$ & $12.8$ \\
MolmoAct2~\citep{fang2026molmoact2} & $47.2$ & $0.0$ & $29.0$ & $12.0$ & $23.0$ & $0.4$ & $18.6$ \\
\capx~\citep{fu2026capx} & $22$ & $18$ & $26$ & $17$ & $12$ & $14$ & $18.2$ \\
\capx (\rats eval)~\citep{rats2026playful} & $27.0$ & $31.0$ & $29.0$ & $16.0$ & $13.0$ & $23.0$ & $23.2$ \\
\rats~\citep{rats2026playful} & $61.0$ & $63.0$ & $43.0$ & $36.0$ & $29.0$ & $31.0$ & $43.8$ \\
\midrule
\spark, \rats protocol (\textbf{Ours}) & $\mathbf{87.8}$ & $\mathbf{96.0}$ & $\mathbf{68.6}$ & $\mathbf{52.2}$ & $\underline{78.0}$ & $\mathbf{84.2}$ & $\mathbf{77.8}$ \\
\spark, strict (\textbf{Ours}) & $\underline{84.0}$ & $\underline{94.6}$ & $\underline{62.2}$ & $\underline{44.2}$ & $\mathbf{78.8}$ & $\underline{82.8}$ & $\underline{74.4}$ \\
\bottomrule
\end{tabular}
\end{table*}

Under the \rats protocol \spark succeeds in $77.8\%$ of trials on average across the six \libpro settings, $34.0$ points above \rats at $43.8\%$ (Table~\ref{tab:libero_main}). Under the strict protocol \spark reaches $74.4\%$, $30.6$ points above \rats. On the same trials the success check after each attempt adds $3.4$ points ($[0.9, 6.4]$, tasks resampled). Per suite we get $91.9\%$ on object, $81.1\%$ on spatial and $60.4\%$ on goal under the \rats protocol, and $89.3\%$, $80.8\%$ and $53.2\%$ under the strict protocol.
Under the \rats protocol \spark gets an average of $69.7\%$ on the language and object perturbations of \libpro, which we did not use to create our grounding rules (Appendix~\S\ref{sec:appendix:provenance}). Other systems that outperform us on \libpro either practice or search for a program per task on training seeds of the evaluation tasks or take several minutes of reasoning per episode (\S\ref{sec:related}). The \capx and \rats rows are comparisons at the system level, because these systems differ from \spark in skill library, language model and controller. The matched arms of \S\ref{sec:libero_dyn_main} hold these parts fixed. Per-task results, with the goal-suite tasks that carry most of the losses, are in Appendix~\S\ref{sec:appendix:provenance}.

\spark was also tested on the \capbench suite of 7 tasks. \spark achieves a 65.1\% success rate with the Qwen planner, compared to \capx's ensemble of three models which achieves 68.3\% (Appendix~\S\ref{sec:appendix:capbench}, Table~\ref{tab:capbench}). We also test if our planner can solve a task without a dedicated skill. When using only base primitives we can stack 10/10 times and wipe 0/20 times (Appendix~\S\ref{sec:appendix:emergence}, Table~\ref{tab:emergence}). We need our wipe skill to apply force and cover the area. Restricting the skill library to the 5 base primitives leads to $424$ of $600$ successes on \libpro compared to $464$ with the full library ($-6.7$ points, $[-13.5, -1.0]$; Appendix~\S\ref{sec:appendix:skill_library}), with the loss in the goal suite.

\subsection{Recovering When Objects Move}
\label{sec:libero_dyn_main}

Scene-diff verification and layered recovery allow us to maintain success when objects are displaced without creating a new plan. All other factors (plan source, skills, executor, controller and perception) are held fixed. Each recovery mechanism is applied alone to all 10 object-suite tasks. The pick target is displaced between $0$ and $10$\,cm along 40 simulator steps in 3 places during an episode, from 5 initial states and on the same seeded schedule per arm (yielding 150 trials per cell). All recoveries rely on the robot's own checks. For scene diff, we re-detect at the onset and during each keypoint move, and check against the detection each label currently references. If displacement is $<10$\,cm, we update the keypoint target in place; if not, we abort the attempt. On a failed attempt we perform layered recovery (the ladder): reassign detections to labels, local re-attempts and finally replanning (retract arm, detect and ask the planner for a new plan; up to 2x). Scripted plans lack any checks and get no replan tier, whereas plans generated by our planner include checks for grasping/placing, with local re-attempts in all arms, including off.

\begin{table}[t]
\centering
\caption{Successes of $150$ under object displacement during execution, strict protocol (10 object-suite tasks, 3 displacement phases, 5 initial states). Within a column group the arms run the same displaced trials and differ only in the recovery mechanisms switched on.}
\label{tab:dyn_main}
\small
\setlength{\tabcolsep}{3.5pt}
\begin{tabular}{lcccc@{\hspace{10pt}}cccc}
\toprule
& \multicolumn{4}{c}{Scripted plans} & \multicolumn{4}{c}{Planner-written plans} \\
\cmidrule(lr){2-5}\cmidrule(lr){6-9}
Recovery & $0$ & $2$ & $5$ & $10$\,cm & $0$ & $2$ & $5$ & $10$\,cm \\
\midrule
Off & $117$ & $87$ & $14$ & $3$ & $120$ & $104$ & $72$ & $71$ \\
Scene diff & $105$ & $82$ & $43$ & $8$ & $123$ & $101$ & $80$ & $73$ \\
Ladder & $120$ & $108$ & $79$ & $88$ & $120$ & $109$ & $85$ & $90$ \\
Scene diff + ladder & $123$ & $101$ & $84$ & $90$ & $\mathbf{123}$ & $\mathbf{109}$ & $\mathbf{99}$ & $\mathbf{107}$ \\
\bottomrule
\end{tabular}
\end{table}

For $10$\,cm displacement, the ladder raises success vs off from $3$ to $88$ of $150$ using scripted plans ($+56.7$ points, $[32.7, 78.7]$, tasks resampled), and from $71$ to $90$ for the planner's plans ($+12.7$, $[4.7, 22.0]$; see Table~\ref{tab:dyn_main}). Scene diff alone yields only marginal improvement at $10$\,cm ($+3.3$, $[0.7, 6.7]$ for scripted; $+1.3$, $[-11.3, 14.7]$ for the planner's plans). Using the planner's plans, scene diff + ladder succeeds in $107$ of $150$ trials ($71.3\%$) at $10$\,cm and $99$ at $5$\,cm. Against a control that requests a new plan after each failure on the same displaced trials (Table~\ref{tab:dyn_matched}), the ladder yields $91$ of $150$ successes at $10$\,cm vs $72$ ($+12.7$ points, $[0.0, 32.0]$). Without displacement they perform similarly ($457$ vs $464$ of $600$), with fewer planner requests for the ladder (Appendix~\S\ref{sec:libero_dyn}).
Two tasks stay at $0$--$2$ of $15$ at $10$\,cm in every arm because a look-alike object is bound to the label. Results for instantaneously displacing objects, including comparisons to VLAs, are in Appendix~\S\ref{sec:libero_dyn}, Table~\ref{tab:libero_dyn}, with detection and recovery times.

\subsection{When Additional Perception Helps}
\label{sec:matched_budget}

\paragraph{Alternative detection descriptions.}
Using alternative detection prompts can increase success on spatial and object suites, but can also decrease success on goal-task perturbations. Detection will either have one description per object or multiple alternatives created once at the start of each trial. A perception fault sweep is in Appendix~\S\ref{sec:appendix:faults}.

\begin{table}[h]
\centering
\caption{Effect of alternative object descriptions on task success ($\%$). ``Single description'' uses one prompt per object from the task language. ``Alternative descriptions'' adds prompt variants. Object and Spatial average position and task perturbations. Goal-task reports the task perturbation setting only. Overall averages all six settings, including the goal-position result in Appendix~\S\ref{sec:appendix:ablation}. Each underlying setting includes ten tasks with $50$ trials per task. Differences are in percentage points.}
\label{tab:perception_main}
\small
\setlength{\tabcolsep}{4pt}
\begin{tabular}{lcccc}
\toprule
Config & Object mean & Spatial mean & Goal-task & Overall \\
\midrule
Single description & $29.9$ & $36.5$ & $22.4$ & $31.9$ \\
Alternative descriptions & $38.1$ & $63.0$ & $14.0$ & $42.7$ \\
\midrule
$\Delta$ & $+8.2$ & $+26.5$ & $-8.4$ & $+10.8$ \\
\bottomrule
\end{tabular}
\end{table}

\subsection{Physical Experiments}
\label{sec:realworld}

\begin{figure*}[t]
\centering
\includegraphics[width=0.88\textwidth]{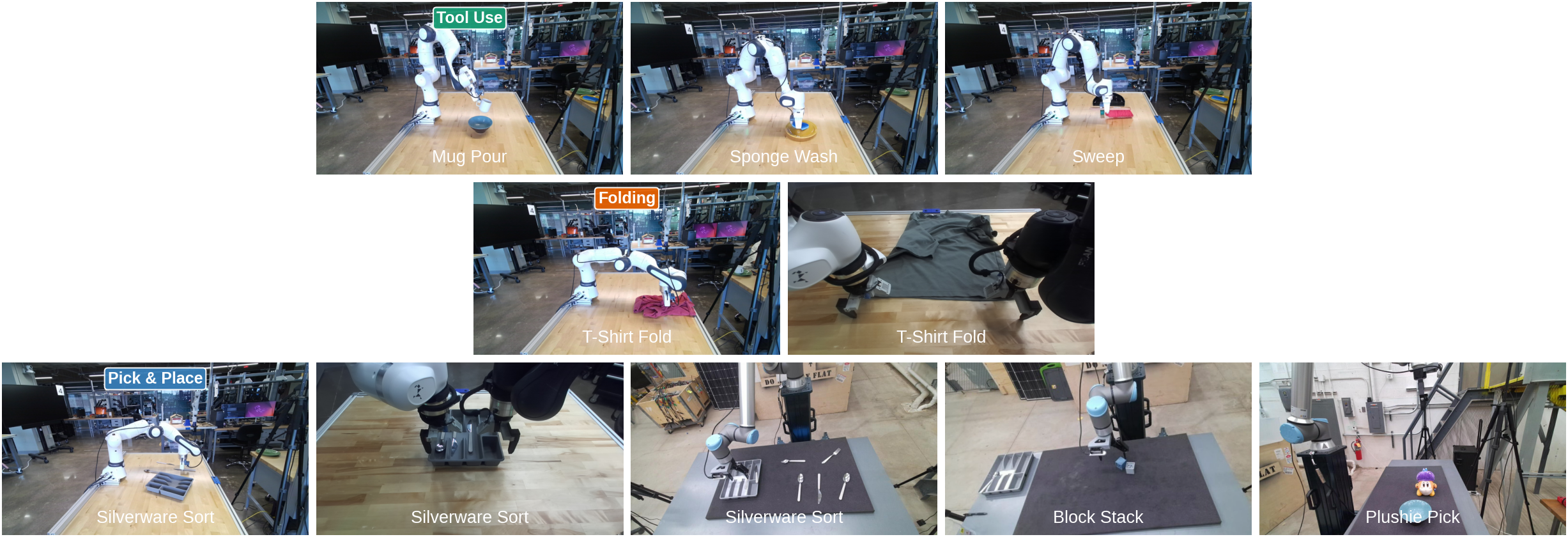}
\caption{Examples of tool use, folding, and object manipulation on UR10e, Franka FR3, and bimanual Franka platforms. Appendix~\S\ref{sec:appendix:physical} reports success rates for all eleven combinations.}
\label{fig:task_grid}
\end{figure*}

We run \spark on the UR10e, Franka FR3 and Bimanual Franka with the same BT grammar and SAM3 pipeline, and no trained models specific to each task. The UR10e rows use the Qwen planner of Table~\ref{tab:libero_main}, and the FR3 and bimanual rows use Gemini 3.5 Flash. All runs on hardware use hardware depth and one description for perception. Over 11 task/platform combinations with 20 randomized trials each \spark was able to complete 165 out of 220 trials (75.0\%, Appendix~\S\ref{sec:appendix:physical}). Some example tasks are shown in Figure~\ref{fig:task_grid}. Each platform uses its own IK solver, gripper interface, workspace limits and contact-control parameters. We run the utensils in tray task on two platforms. UR10e completes 20/20 trials and FR3 completes 16/20 trials. On the Bimanual Franka platform we can fold a sleeve in 15/20 trials and complete all three folds in 6/20 trials. With objects moved by hand during the episode (Table~\ref{tab:physical_live}), the UR10e completes $10/10$ Utensils in tray, Stack blocks and Plushie in bowl trials and $8/10$ Utensils in bowl trials, with $36$ of the $40$ on a single plan.

\section{Discussion}
\label{sec:discussion}

If an object gets displaced, then we can have a valid action sequence but no longer know where to perform the action. \spark explicitly accounts for this by resolving object labels during execution and attempting to recover within the existing behavior tree. With identical scripted plans the ladder raises success under $10$\,cm displacements from $3$ to $88$ of $150$, and with planner-written plans the ladder does at least as well as asking for a new plan after each failure, with fewer planner requests (Table~\ref{tab:dyn_main}, Table~\ref{tab:dyn_matched}). Some unsuccessful executions can be fixed by correcting where the action is performed while keeping the action sequence.

Being able to keep a plan connected to the scene also depends on being able to observe the relevant objects and having skills to complete those actions. Adding alternate detection descriptions gave a $26.5$ point increase on average in the spatial suite and an $8.4$ point decrease on the goal tasks (Table~\ref{tab:perception_main}, Appendix~\S\ref{sec:appendix:ablation}). Our skill-emergence ablation tests which skills the planner can build by itself. Our planner is able to stack with the base skills in all 10 trials but fails to wipe in all 20 trials without the dedicated skill (Appendix~\S\ref{sec:appendix:emergence}, Table~\ref{tab:emergence}). Some skills can be represented by sequences of our current skills, but contact regulation and full coverage need to be in the executor. What we leave to future work is knowing how execution feedback should pick an additional observation, plan or skill.

\paragraph{Rollouts as labelled data.}
Every step \spark executes is a typed node with a checked outcome, so each rollout is recorded as labelled segments: the tree, a record per step with its target detection, times and outcome, the trajectory with the active step, video, and a pass/fail label. The runs of Table~\ref{tab:libero_main} hold $6{,}000$ trials with a success label, $7{,}696$ executed plans, $13{,}186$ check events and $14{,}081$ grasp and placement steps with a recorded outcome. On the robot, successful trees are stored and can be reused as planner context (\S\ref{sec:planning}), and the executor can reset a scene by running a placement in reverse, so collection can run unattended (Appendix~\S\ref{sec:appendix:physical}).

\paragraph{Limitations.}
\label{sec:limitations}
Our \libpro evaluation reads no ground-truth object poses from the simulator and does not practice on the evaluated tasks, although whether this is a limitation is up for debate. Our goal suite has many articulated objects and our score is lowest here ($60.4\%$ under the \rats protocol and $53.2\%$ strict). We used the tasks in \lib to help develop our controllers and skills. Local recovery assumes that the plan is still valid. In cases where the plan itself is wrong, a new plan is required; we did not test whether local recovery or immediate replanning is preferable there.
Our simulated displacement experiments follow a scripted schedule, and so we cannot claim reliable grasping of continuously moving objects. Additionally, real-world robots require per-platform tuning; we evaluate hardware over a fixed set of tasks with 20 trials each. This will be extended to test generalization on a wide range of physical skills at scale.

\section{Conclusion}
\label{sec:conclusion}

\spark allows us to separate our high level plan of how to manipulate objects from their current locations through generating a behavior tree over a fixed set of skills we can execute. The executor will associate these with the current objects and check grasps and placements where the plan includes those checks, attempting local recovery if possible, and if not, it will query a new plan.
With a fixed plan we can utilize scene-diff verification and layered recovery to reach $60.0\%$ success rates with $10$\,cm dynamic displacements. We reached an average of $77.8\%$ success on all 6 position and task settings of \libpro without training or practicing on any of the tasks being tested ($74.4\%$ when only the robot's own checks decide retries). We also reached a $75\%$ success rate on 11 physical tasks across three robot embodiments with platform specific control tuning. We think that local recovery from mistakes within a valid plan should be an explicit part of developing robust and interpretable manipulation systems. There is more work to be done to allow verification from onboard sensors, improve contact and articulated control, and allow execution feedback to decide between updating perception, local correction and a new plan, we leave this to future work.

\bibliography{references}

@article{zhou2025liberopro,
  title   = {{LIBERO-PRO}: Towards Robust and Fair Evaluation of Vision-Language-Action Models Beyond Memorization},
  author  = {Zhou, Xueyang and Xu, Yangming and Tie, Guiyao and Chen, Yongchao and Zhang, Guowen and Chu, Duanfeng and Zhou, Pan and Sun, Lichao},
  journal = {arXiv preprint arXiv:2510.03827},
  year    = {2025}
}

@inproceedings{liu2023libero,
  title={{LIBERO}: Benchmarking Knowledge Transfer for Lifelong Robot Learning},
  author={Liu, Bo and Zhu, Yifeng and Gao, Chongkai and Feng, Yihao and Liu, Qiang and Zhu, Yuke and Stone, Peter},
  booktitle={Advances in Neural Information Processing Systems (NeurIPS), Datasets and Benchmarks Track},
  year={2023},
}

@inproceedings{kim2024openvla,
  title={{OpenVLA}: An Open-Source Vision-Language-Action Model},
  author={Kim, Moo Jin and Pertsch, Karl and Karamcheti, Siddharth and Xiao, Ted and Balakrishna, Ashwin and Nair, Suraj and Rafailov, Rafael and Foster, Ethan P. and Sanketi, Pannag R. and Vuong, Quan and Kollar, Thomas and Burchfiel, Benjamin and Tedrake, Russ and Sadigh, Dorsa and Levine, Sergey and Liang, Percy and Finn, Chelsea},
  booktitle={Proceedings of the 8th Conference on Robot Learning (CoRL)},
  series={Proceedings of Machine Learning Research},
  volume={270},
  pages={2679--2713},
  publisher={PMLR},
  year={2024},
}

@inproceedings{black2024pi0,
  title={{$\pi_0$}: A Vision-Language-Action Flow Model for General Robot Control},
  author={Black, Kevin and Brown, Noah and Driess, Danny and Esmail, Adnan and Equi, Michael Robert and Finn, Chelsea and Fusai, Niccolo and Groom, Lachy and Hausman, Karol and Ichter, Brian and Jakubczak, Szymon and Jones, Tim and Ke, Liyiming and Levine, Sergey and Li-Bell, Adrian and Mothukuri, Mohith and Nair, Suraj and Pertsch, Karl and Shi, Lucy Xiaoyang and Smith, Laura and Tanner, James and Vuong, Quan and Walling, Anna and Wang, Haohuan and Zhilinsky, Ury},
  booktitle={Proceedings of Robotics: Science and Systems (RSS)},
  year={2025},
  doi={10.15607/RSS.2025.XXI.010},
}

@inproceedings{physicalintelligence2025pi05,
  title     = {{$\pi_{0.5}$}: a Vision-Language-Action Model with Open-World Generalization},
  author    = {Black, Kevin and Brown, Noah and Darpinian, James and Dhabalia, Karan and Driess, Danny and Esmail, Adnan and Equi, Michael Robert and Finn, Chelsea and Fusai, Niccolo and Galliker, Manuel Y. and Ghosh, Dibya and Groom, Lachy and Hausman, Karol and Ichter, Brian and Jakubczak, Szymon and Jones, Tim and Ke, Liyiming and LeBlanc, Devin and Levine, Sergey and Li-Bell, Adrian and Mothukuri, Mohith and Nair, Suraj and Pertsch, Karl and Ren, Allen Z. and Shi, Lucy Xiaoyang and Smith, Laura and Springenberg, Jost Tobias and Stachowicz, Kyle and Tanner, James and Vuong, Quan and Walke, Homer and Walling, Anna and Wang, Haohuan and Yu, Lili and Zhilinsky, Ury},
  booktitle = {Proceedings of The 9th Conference on Robot Learning (CoRL)},
  series    = {Proceedings of Machine Learning Research},
  volume    = {305},
  pages     = {17--40},
  publisher = {PMLR},
  year      = {2025}
}

@inproceedings{fu2026capx,
  title={{CaP-X}: A Framework for Benchmarking and Improving Coding Agents for Robot Manipulation},
  author={Fu, Letian and Yu, Justin and El-Refai, Karim and Kou, Ethan and Xue, Haoru and Huang, Huang and Xiao, Wenli and Fei-Fei, Li and Shi, Guanya and Wu, Jiajun and Sastry, S. Shankar and Zhu, Yuke and Goldberg, Ken and Fan, Linxi},
  booktitle={Proceedings of the 43rd International Conference on Machine Learning (ICML)},
  series={Proceedings of Machine Learning Research},
  volume={306},
  pages={32155--32212},
  publisher={PMLR},
  year={2026},
}

@article{xiao2026enpire,
  title   = {{ENPIRE}: Agentic Robot Policy Self-Improvement in the Real World},
  author  = {Xiao, Wenli and Xie, Jia and Zhang, Tonghe and Lin, Haotian and Fu, Letian and Xue, Haoru and Lu, Jalen and Yang, Yi and Dai, Cunxi and Wang, Zi and Wu, Jimmy and Wang, Guanzhi and Sastry, S. Shankar and Goldberg, Ken and Fan, Linxi and Zhu, Yuke and Shi, Guanya},
  journal = {arXiv preprint arXiv:2606.19980},
  year    = {2026}
}

@inproceedings{grant2026sae,
  title={Not All Features Are Created Equal: A Mechanistic Study of Vision-Language-Action Models},
  author={Grant, Bryce and Zhao, Xijia and Wang, Peng},
  booktitle={Advances in Neural Information Processing Systems (NeurIPS)},
  year={2026},
  note={arXiv:2603.19233},
}

@inproceedings{liang2023codepolicies,
  title={Code as Policies: Language Model Programs for Embodied Control},
  author={Liang, Jacky and Huang, Wenlong and Xia, Fei and Xu, Peng and Hausman, Karol and Ichter, Brian and Florence, Pete and Zeng, Andy},
  booktitle={IEEE International Conference on Robotics and Automation (ICRA)},
  pages={9493--9500},
  year={2023},
  doi={10.1109/ICRA48891.2023.10160591},
}

@inproceedings{ahn2022saycan,
  title={Do As I Can, Not As I Say: Grounding Language in Robotic Affordances},
  author={Ichter, Brian and Brohan, Anthony and Chebotar, Yevgen and Finn, Chelsea and Hausman, Karol and Herzog, Alexander and Ho, Daniel and Ibarz, Julian and Irpan, Alex and Jang, Eric and Julian, Ryan and Kalashnikov, Dmitry and Levine, Sergey and Lu, Yao and Parada, Carolina and Rao, Kanishka and Sermanet, Pierre and Toshev, Alexander T. and Vanhoucke, Vincent and Xia, Fei and Xiao, Ted and Xu, Peng and Yan, Mengyuan and Brown, Noah and Ahn, Michael and Cortes, Omar and Sievers, Nicolas and Tan, Clayton and Xu, Sichun and Reyes, Diego and Rettinghouse, Jarek and Quiambao, Jornell and Pastor, Peter and Luu, Linda and Lee, Kuang-Huei and Kuang, Yuheng and Jesmonth, Sally and Joshi, Nikhil J. and Jeffrey, Kyle and Ruano, Rosario Jauregui and Hsu, Jasmine and Gopalakrishnan, Keerthana and David, Byron and Zeng, Andy and Fu, Chuyuan Kelly},
  booktitle={Proceedings of the 6th Conference on Robot Learning (CoRL)},
  series={Proceedings of Machine Learning Research},
  volume={205},
  pages={287--318},
  publisher={PMLR},
  year={2022},
}

@book{colledanchise2018bt,
  title={Behavior Trees in Robotics and {AI}: An Introduction},
  author={Colledanchise, Michele and {\"O}gren, Petter},
  publisher={CRC Press},
  address={Boca Raton},
  year={2018},
  doi={10.1201/9780429489105},
}

@inproceedings{wang2023selfconsistency,
  title     = {Self-Consistency Improves Chain of Thought Reasoning in Language Models},
  author    = {Wang, Xuezhi and Wei, Jason and Schuurmans, Dale and Le, Quoc and Chi, Ed H. and Narang, Sharan and Chowdhery, Aakanksha and Zhou, Denny},
  booktitle = {International Conference on Learning Representations (ICLR)},
  year      = {2023}
}

@inproceedings{carion2025sam3,
  title={{SAM 3}: Segment Anything with Concepts},
  author={Carion, Nicolas and Gustafson, Laura and Hu, Yuan-Ting and Debnath, Shoubhik and Hu, Ronghang and Suris, Didac and Ryali, Chaitanya and Alwala, Kalyan Vasudev and Khedr, Haitham and Huang, Andrew and Lei, Jie and Ma, Tengyu and Guo, Baishan and Kalla, Arpit and Marks, Markus and Greer, Joseph and Wang, Meng and Sun, Peize and R{\"a}dle, Roman and Afouras, Triantafyllos and Mavroudi, Effrosyni and Xu, Katherine and Wu, Tsung-Han and Zhou, Yu and Momeni, Liliane and Hazra, Rishi and Ding, Shuangrui and Vaze, Sagar and Porcher, Francois and Li, Feng and Li, Siyuan and Kamath, Aishwarya and Cheng, Ho Kei and Doll{\'a}r, Piotr and Ravi, Nikhila and Saenko, Kate and Zhang, Pengchuan and Feichtenhofer, Christoph},
  booktitle={International Conference on Learning Representations (ICLR)},
  year={2026},
}

@article{zhu2020robosuite,
  title   = {robosuite: A Modular Simulation Framework and Benchmark for Robot Learning},
  author  = {Zhu, Yuke and Wong, Josiah and Mandlekar, Ajay and Mart{\'i}n-Mart{\'i}n, Roberto and Joshi, Abhishek and Lin, Kevin and Maddukuri, Abhiram and Nasiriany, Soroush and Zhu, Yifeng},
  journal = {arXiv preprint arXiv:2009.12293},
  year    = {2020}
}

@inproceedings{kim2025pyroki,
  title={{PyRoki}: A Modular Toolkit for Robot Kinematic Optimization},
  author={Kim, Chung Min and Yi, Brent and Choi, Hongsuk and Ma, Yi and Goldberg, Ken and Kanazawa, Angjoo},
  booktitle={IEEE/RSJ International Conference on Intelligent Robots and Systems (IROS)},
  year={2025},
  doi={10.1109/IROS60139.2025.11246651},
}

@inproceedings{lim2024equigraspflow,
  title={{EquiGraspFlow}: {SE(3)}-Equivariant 6-{DoF} Grasp Pose Generative Flows},
  author={Lim, Byeongdo and Kim, Jongmin and Kim, Jihwan and Lee, Yonghyeon and Park, Frank C.},
  booktitle={Proceedings of the 8th Conference on Robot Learning (CoRL)},
  series={Proceedings of Machine Learning Research},
  volume={270},
  pages={5067--5086},
  year={2024},
  publisher={PMLR},
}

@inproceedings{fang2026molmoact2,
  title={{MolmoAct2}: Action Reasoning Models for Real-World Deployment},
  author={Fang, Haoquan and Duan, Jiafei and Clay, Donovan and Wang, Sam and Liu, Shuo and Huang, Weikai and Fan, Xiang and Tsai, Wei-Chuan and Chen, Shirui and Wang, Yi Ru and Xing, Shanli and Cho, Jaemin and Park, Jae Sung and Eftekhar, Ainaz and Sushko, Peter and Farley, Karen and Wadhwa, Angad and Harrison, Cole and Han, Winson and Lee, Ying-Chun and VanderBilt, Eli and Hendrix, Rose and Ellawela, Suveen and Ngoo, Lucas and Chai, Joyce and Ren, Zhongzheng and Farhadi, Ali and Fox, Dieter and Krishna, Ranjay},
  booktitle={Conference on Robot Learning (CoRL)},
  year={2026},
  note={arXiv:2605.02881},
}

@article{vlmbtfailure2025,
  title   = {A Unified Framework for Real-Time Failure Handling in
             Robotics Using {VLM}s, Reactive Planner and Behavior Trees},
  author  = {Ahmad, Faseeh and Ismail, Hashim and Styrud, Jonathan and others},
  journal = {arXiv preprint arXiv:2503.15202},
  year    = {2025},
  note    = {IEEE CASE 2025}
}

@inproceedings{singh2023progprompt,
  title     = {{ProgPrompt}: Generating Situated Robot Task Plans using Large Language Models},
  author    = {Singh, Ishika and Blukis, Valts and Mousavian, Arsalan and Goyal, Ankit and Xu, Danfei and Tremblay, Jonathan and Fox, Dieter and Thomason, Jesse and Garg, Animesh},
  booktitle = {IEEE International Conference on Robotics and Automation (ICRA)},
  pages     = {11523--11530},
  doi       = {10.1109/ICRA48891.2023.10161317},
  year      = {2023}
}

@inproceedings{kwok2025robomonkey,
  title={{RoboMonkey}: Scaling Test-Time Sampling and Verification for Vision-Language-Action Models},
  author={Kwok, Jacky and Agia, Christopher and Sinha, Rohan and Foutter, Matt and Li, Shulu and Stoica, Ion and Mirhoseini, Azalia and Pavone, Marco},
  booktitle={Proceedings of the 9th Conference on Robot Learning (CoRL)},
  series={Proceedings of Machine Learning Research},
  volume={305},
  pages={3200--3217},
  publisher={PMLR},
  year={2025},
}

@inproceedings{chen2024llmobtea,
  author    = {Chen, Xinglin and Cai, Yishuai and Mao, Yunxin and Li, Minglong and Yang, Wenjing and Xu, Weixia and Wang, Ji},
  title     = {Integrating Intent Understanding and Optimal Behavior Planning for Behavior Tree Generation from Human Instructions},
  booktitle = {Proceedings of the 33rd International Joint Conference on Artificial Intelligence (IJCAI)},
  year      = {2024},
  pages     = {6832--6840},
  doi       = {10.24963/ijcai.2024/755}
}

@inproceedings{huang2024rekep,
  title={{ReKep}: Spatio-Temporal Reasoning of Relational Keypoint Constraints for Robotic Manipulation},
  author={Huang, Wenlong and Wang, Chen and Li, Yunzhu and Zhang, Ruohan and Fei-Fei, Li},
  booktitle={Proceedings of the 8th Conference on Robot Learning (CoRL)},
  series={Proceedings of Machine Learning Research},
  volume={270},
  pages={4573--4602},
  publisher={PMLR},
  year={2024},
}

@inproceedings{rats2026playful,
  title     = {Playful Agentic Robot Learning},
  author    = {Zhang, Junyi and Ge, Jiaxin and Yoo, Hanjun and Fu, Letian and Yang, Zihan and Liu, Yaowei and Saravanan, Raj and Yin, Shaofeng and Yu, Justin and Niu, Dantong and Wang, Zirui and Herzig, Roei and Goldberg, Ken and Bai, Yutong and Chan, David M. and Stoica, Ion and Kanazawa, Angjoo and Lei, Jiahui and Feng, Haiwen and Darrell, Trevor},
  booktitle = {Conference on Robot Learning (CoRL)},
  year      = {2026}
}

@article{lu2026aspire,
  title   = {{ASPIRE}: Agentic {/Skills} Discovery for Robotics},
  author  = {Lu, Runyu and Wu, Yubo and Kou, Ethan and Fu, Letian and Xiao, Wenli and Mandlekar, Ajay and Xu, Yinzhen and Shi, Guanya and Goldberg, Ken and Chen, Ang and Chowdhury, Mosharaf and Zhu, Yuke and Fan, Linxi and Wang, Guanzhi},
  journal = {arXiv preprint arXiv:2607.00272},
  year    = {2026}
}

@article{salfity2026contractbt,
  title   = {Contract-Grounded Behavior Tree Synthesis via Coding Agent},
  author  = {Salfity, Jonathan and Anderson, Robert Blake and Pryor, Mitch},
  journal = {arXiv preprint arXiv:2607.12220},
  year    = {2026}
}

@inproceedings{jh2025ulag,
  title     = {{U-LAG}: Uncertainty-Aware, Lag-Adaptive Goal Retargeting for Robotic Manipulation},
  author    = {{Anamika J H} and Muraleedharan, Anujith},
  booktitle = {IROS 2025 Workshop on Perception and Planning for Mobile Manipulation in Changing Environments},
  year      = {2025},
  note      = {arXiv:2510.02526}
}

@inproceedings{pumacay2024colosseum,
  title     = {{THE COLOSSEUM}: A Benchmark for Evaluating Generalization for Robotic Manipulation},
  author    = {Pumacay, Wilbert and Singh, Ishika and Duan, Jiafei and Krishna, Ranjay and Thomason, Jesse and Fox, Dieter},
  booktitle = {Robotics: Science and Systems (RSS)},
  year      = {2024}
}

@article{zhang2026harness,
  title   = {Harness {VLA}: Steering Frozen {VLAs} into Reliable Manipulation Primitives via Memory-Guided Agents},
  author  = {Zhang, Yixian and Zhang, Huanming and Gao, Feng and Li, Xiao and Liu, Zhihao and Nie, Yi and Zhu, Chunyang and Qiu, Jiaxing and Yan, Yuchen and Liu, Jiyuan and Tang, Wenhao and Rao, Jiaji and Fang, Zhengru and Wei, Changxu and Wang, Yu and Ding, Wenbo and Yu, Chao},
  journal = {arXiv preprint arXiv:2607.08448},
  year    = {2026}
}

@article{zetta2026,
  title   = {Zetta {$\zeta$}: An Efficient Closed-Loop Embodied Harness for Self-Evolving Physical Intelligence},
  author  = {Ding, Xin and Mi, Liang and Huang, Mingzhe and Wang, Zixuan and Zhang, Chao and Hao, Zixu and Chen, Fu and Li, Xiangyu and Zheng, Yikai and Guo, Yaoyu and Wang, Weijun and Li, Kun and Wu, Hao and Liu, Yunxin and Cao, Ting},
  journal = {arXiv preprint arXiv:2608.16590},
  year    = {2026}
}

@inproceedings{guo2024doremi,
  title     = {{DoReMi}: Grounding Language Model by Detecting and Recovering from Plan-Execution Misalignment},
  author    = {Guo, Yanjiang and Wang, Yen-Jen and Zha, Lihan and Chen, Jianyu},
  booktitle = {IEEE/RSJ International Conference on Intelligent Robots and Systems (IROS)},
  year      = {2024},
  pages     = {12124--12131},
  note      = {arXiv:2307.00329}
}

@inproceedings{zhou2025codeasmonitor,
  title     = {Code-as-Monitor: Constraint-aware Visual Programming for Reactive and Proactive Robotic Failure Detection},
  author    = {Zhou, Enshen and Su, Qi and Chi, Cheng and Zhang, Zhizheng and Wang, Zhongyuan and Huang, Tiejun and Sheng, Lu and Wang, He},
  booktitle = {IEEE/CVF Conference on Computer Vision and Pattern Recognition (CVPR)},
  pages     = {6919--6929},
  year      = {2025}
}

@article{yin2026core,
  title   = {Imagining Recovery: Inference-Time Counterfactual Realignment for Vision-Language-Action Models},
  author  = {Zhang, Yanyan and Liu, Disheng and Ye, Kai and Song, Chaoda and Li, Xinpeng and Hariri, Mohsen and Singh, Vikash and Yin, Yu and Chaudhary, Vipin},
  journal = {arXiv preprint arXiv:2608.14822},
  year    = {2026}
}

@article{song2026embodiedrsi,
  title   = {{EmbodiedRSI}: Active Continual Robot Learning Through Hypothesis-Guided Co-Evolution},
  author  = {Song, Python and Liang, Zhixuan and Fu, Kelsey and Wang, Mengdi and Yang, Junfeng and Liu, Shilong},
  journal = {arXiv preprint arXiv:2610.10498},
  year    = {2026}
}

@article{wang2026rpg,
  title   = {Reconstruct, Practice, Go Real: Guided Self-Improvement for Embodied Agents},
  author  = {Wang, Yen-Jen and Jiang, Haozhe and Deng, Shuying and Xue, Haoru and Ye, Weirui and Duan, Rocky and Haghtalab, Nika and Sastry, S. Shankar and Abbeel, Pieter and Qi, Haozhi},
  journal = {arXiv preprint arXiv:2610.02204},
  year    = {2026}
}

@article{chu2026openrua,
  title   = {{OpenRUA}: Robot-Use Agents Are Zero-Shot Visuomotor Policies},
  author  = {Chu, Zhaoyang and Barr, Earl T. and Le Goues, Claire and O'Hearn, Peter and Harman, Mark and Sarro, Federica and Ye, He},
  journal = {arXiv preprint arXiv:2610.02459},
  year    = {2026}
}

@misc{yang2026ora,
  title        = {{ORA}: Open Robot-Use Agents},
  author       = {Yang, Yuanbo and Chen, Siyi and Gao, Jun and Zhang, Hao},
  year         = {2026},
  month        = oct,
  howpublished = {Hao AI Lab blog, \url{https://haoailab.com/blogs/ora/}}
}

@inproceedings{kwon2023vllm,
  title     = {Efficient Memory Management for Large Language Model Serving with {PagedAttention}},
  author    = {Kwon, Woosuk and Li, Zhuohan and Zhuang, Siyuan and Sheng, Ying and Zheng, Lianmin and Yu, Cody Hao and Gonzalez, Joseph and Zhang, Hao and Stoica, Ion},
  booktitle = {Proceedings of the 29th Symposium on Operating Systems Principles (SOSP)},
  year      = {2023}
}

@inproceedings{cheng2024autoround,
  title     = {Optimize Weight Rounding via Signed Gradient Descent for the Quantization of {LLM}s},
  author    = {Cheng, Wenhua and Zhang, Weiwei and Shen, Haihao and Cai, Yiyang and He, Xin and Kaokao, Lv and Liu, Yi},
  booktitle = {Findings of the Association for Computational Linguistics: EMNLP 2024},
  pages     = {11332--11350},
  year      = {2024}
}

@article{liu2026roboauto,
  title   = {{RoboAuto}: Autonomous Robot Program Search through Multi-Level Physical Understanding},
  author  = {Liu, Buzhao and Ye, Jianing and Yuan, Xiu and Feng, Yangtong and Huang, Chengsong and Dinh, Tyler and Vorobeychik, Yevgeniy and Zhang, Ning and Zhang, Chongjie},
  journal = {arXiv preprint arXiv:2610.36533},
  year    = {2026}
}

@article{zhang2026liberomax,
  title   = {{LIBERO-MAX}: Do Robot Policies Adapt When the World Changes?},
  author  = {Zhang, Yunbei and Jin, Zijian and Liu, Yuanzhe and Wang, Janet and Zhang, Xilun and Zhang, Yuyou and Zhang, Zhenyu and Zhang, Daoan and Niu, Shuaicheng and Li, Gen and Yang, Jianfei and Hamm, Jihun and Lourentzou, Ismini and Ye, Weirui and Liu, Bo and Stone, Peter and Pavone, Marco},
  journal = {arXiv preprint arXiv:2609.36518},
  year    = {2026}
}

@article{ren2026liberodynamic,
  title   = {{LIBERO-Dynamic}: Benchmarking Robotic Manipulation in Dynamic Environments},
  author  = {Ren, Shaojie and Sun, Jingwen and Liu, Zezhi and Chen, Hongjin and Jin, Xin and Zou, Wei and Chen, Zhibo},
  journal = {arXiv preprint arXiv:2610.24579},
  year    = {2026}
}

@article{zhou2026liberopro2,
  title   = {{LIBERO-Pro2.0}: A Benchmark for Evaluating the Robustness of Embodied Foundation Models},
  author  = {Zhou, Xueyang and Xu, Yangming and Li, Qianjiang and Tie, Guiyao and Wu, Junhao and Huang, Ziheng and Zhang, Guowen and Yuan, Zhijie and Wu, Di and Chu, Duanfeng and Zhou, Pan and Sun, Lichao and Chen, Yongchao},
  journal = {arXiv preprint arXiv:2610.24587},
  year    = {2026}
}

@article{agia2024sentinel,
  title   = {Unpacking Failure Modes of Generative Policies: Runtime Monitoring of Consistency and Progress},
  author  = {Agia, Christopher and Sinha, Rohan and Yang, Jingyun and Cao, Zi-ang and Antonova, Rika and Pavone, Marco and Bohg, Jeannette},
  journal = {arXiv preprint arXiv:2410.04640},
  year    = {2024}
}

@inproceedings{sinha2024aesop,
  title     = {Real-Time Anomaly Detection and Reactive Planning with Large Language Models},
  author    = {Sinha, Rohan and Elhafsi, Amine and Agia, Christopher and Foutter, Matt and Schmerling, Edward and Pavone, Marco},
  booktitle = {Robotics: Science and Systems (RSS)},
  year      = {2024},
  note      = {arXiv:2407.08735}
}

@article{huang2025fail2progress,
  title   = {{Fail2Progress}: Learning from Real-World Robot Failures with Stein Variational Inference},
  author  = {Huang, Yixuan and Alvina, Novella and Shanthi, Mohanraj Devendran and Hermans, Tucker},
  journal = {arXiv preprint arXiv:2509.01746},
  year    = {2025}
}

@article{coradeschi2003anchoring,
  title   = {An Introduction to the Anchoring Problem},
  author  = {Coradeschi, Silvia and Saffiotti, Alessandro},
  journal = {Robotics and Autonomous Systems},
  volume  = {43},
  pages   = {85--96},
  year    = {2003},
  doi     = {10.1016/S0921-8890(03)00021-6}
}

@article{fikes1972planex,
  title   = {Learning and Executing Generalized Robot Plans},
  author  = {Fikes, Richard E. and Hart, Peter E. and Nilsson, Nils J.},
  journal = {Artificial Intelligence},
  volume  = {3},
  pages   = {251--288},
  year    = {1972},
  doi     = {10.1016/0004-3702(72)90051-3}
}

@phdthesis{firby1989rap,
  title  = {Adaptive Execution in Complex Dynamic Worlds},
  author = {Firby, R. James},
  school = {Yale University},
  year   = {1989},
  note   = {Research Report 672}
}

@article{liu2023reflect,
  title   = {{REFLECT}: Summarizing Robot Experiences for Failure Explanation and Correction},
  author  = {Liu, Zeyi and Bahety, Arpit and Song, Shuran},
  journal = {arXiv preprint arXiv:2306.15724},
  year    = {2023}
}

@article{cornelio2024recover,
  title   = {Recover: A Neuro-Symbolic Framework for Failure Detection and Recovery},
  author  = {Cornelio, Cristina and Diab, Mohammed},
  journal = {arXiv preprint arXiv:2404.00756},
  year    = {2024}
}

@article{ren2026cybowaiter,
  title   = {{Cybo-Waiter}: A Physical Agentic Framework for Humanoid Whole-Body Locomotion-Manipulation},
  author  = {Ren, Peng and Ge, Haoyang and Qi, Chuan and Huang, Cong and Li, Hong and Zhao, Jiang and Chi, Pei and Chen, Kai},
  journal = {arXiv preprint arXiv:2603.10675},
  year    = {2026}
}
\bibliographystyle{iclr2027_conference}
\appendix
\appendix
\clearpage

\section{Implementation Details}
\label{sec:appendix:impl}

\begin{figure}[t]
\centering
\begin{minipage}[t]{0.325\linewidth}\centering
\includegraphics[width=\linewidth]{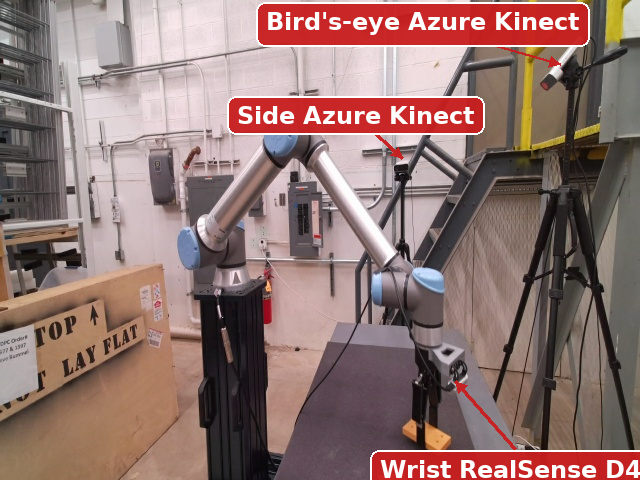}\\[2pt]
{\small (a) UR10e + Robotiq 2F-85}
\end{minipage}\hfill
\begin{minipage}[t]{0.325\linewidth}\centering
\includegraphics[width=\linewidth]{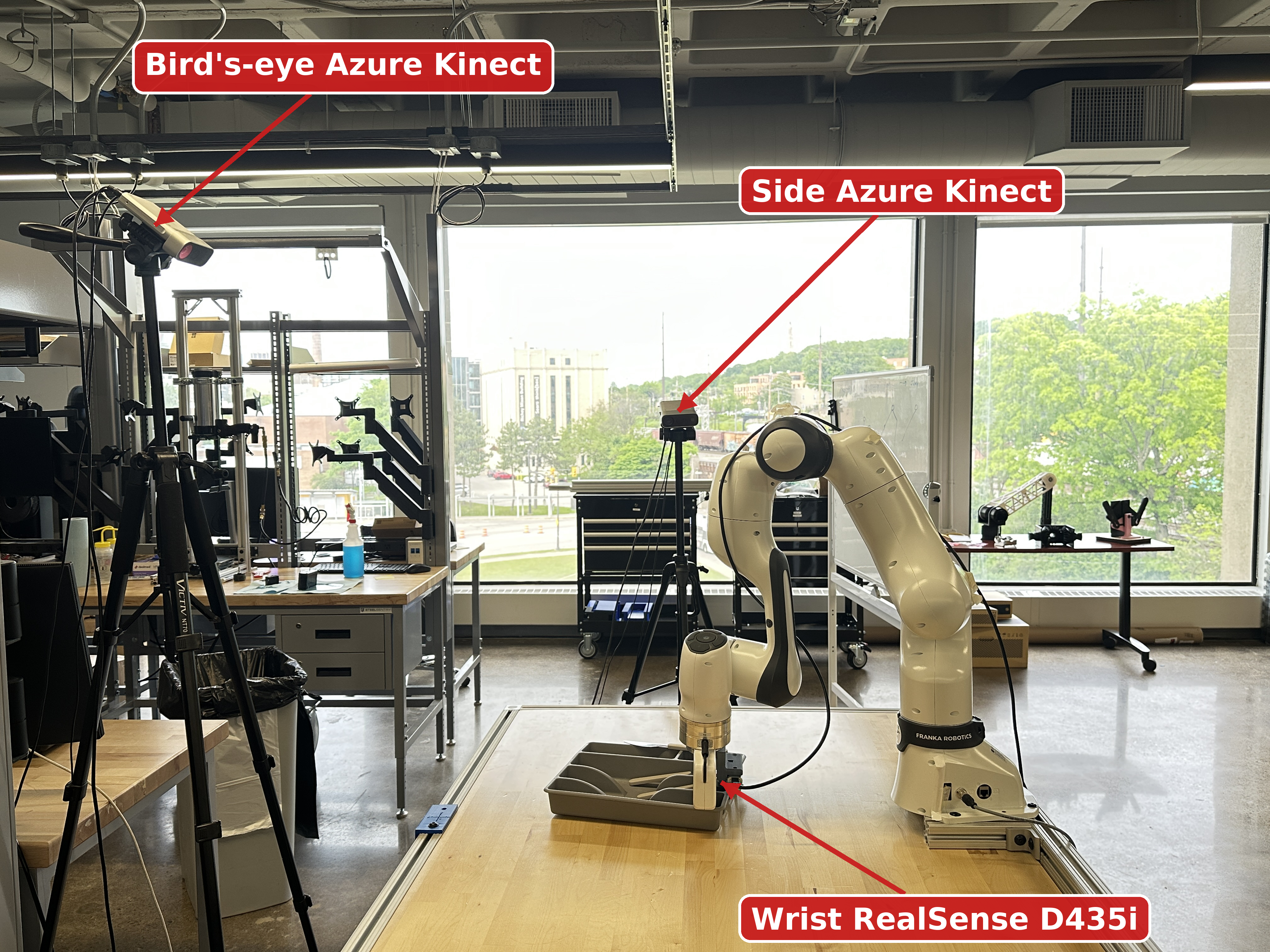}\\[2pt]
{\small (b) FR3 + Franka Hand}
\end{minipage}\hfill
\begin{minipage}[t]{0.325\linewidth}\centering
\includegraphics[width=\linewidth]{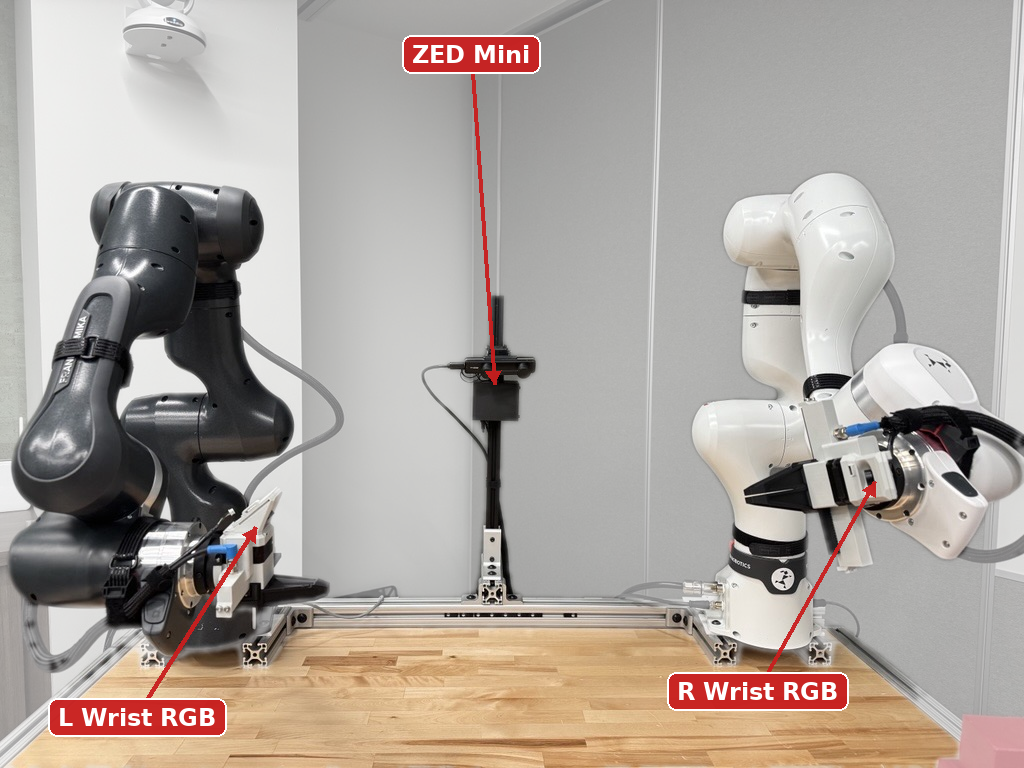}\\[2pt]
{\small (c) Bimanual workcell (Panda left, FR3 right, Source Robotics SSG-48 grippers)}
\end{minipage}
\caption{Rigs used on our three physical platforms. Each image is labelled with the respective camera names. Our single arm setups (a, b) consist of a top down and side Azure Kinect DK along with a RealSense D435i on the wrist. For our bimanual workcell (c) we have one back facing ZED Mini pointed at the workspace and no Azure Kinect DK. Instead there is one OV9732 on each wrist. Our simulated setup is constant for the \libpro environment, which is the default \texttt{agentview} 3rd person and the \texttt{robot0\_eye\_in\_hand} camera. We did not move the cameras.}
\label{fig:setup}
\end{figure}

Here we report the component-level settings we used to reproduce results on \libpro and \capbench. Unless specified below we did not tune on a per-task basis beyond what is reported here. The camera rigs of our three physical platforms are shown in Figure~\ref{fig:setup}.

\subsection{SAM3 Detection}
\label{sec:appendix:sam3}

We use the released checkpoint for SAM3~\citep{carion2025sam3} with its vanilla open-vocabulary text prompted API. We use the built-in confidence threshold for detections. In our adaptive setup described above the variant-selection rule from \S\ref{sec:adaptive} replaces hard thresholding. We keep the objects own name if it is present in SAM3 detections, else we keep the variant with the highest confidence multiplied by 1 if there is one detection, 0.6 if there are two and 0.1 if there are more. Prompts are short noun phrases (1--6 tokens) suggested by our planner model (Qwen3.8-27B from Table~\ref{tab:libero_main} and Gemini in the perception ablation) using the goal objects from the task file as well as a static list of phrases. With the \emph{+BDDL names} setting we additionally query LIBERO's canonical-name dictionary. We query both cameras every perception cycle: the agentview camera ($640\times 480$) and the \texttt{robot0\_eye\_in\_hand} wrist camera ($640\times 480$), with the same prompts.

\subsection{Planner Models (BT Generator and Prompt Synthesiser)}
\label{sec:appendix:gemini}

For simulation our main-table planner is Qwen3.8-27B with 4bit weights (\texttt{MIRALABS/Qwen3.8-27B-W4A16-AutoRound}) served with vllm with thinking disabled, both for the behavior tree call (temperature $0$) as well as the prompt suggestion call. Our ablation configuration uses \texttt{gemini-3.1-pro-preview} with temperature $0$. For the FR3 and bimanual rows of Table~\ref{tab:physical} we use Gemini 3.5 Flash (\texttt{gemini-3.5-flash}) with temperature $0.3$. The UR10e rows and the live-displacement trials of Table~\ref{tab:physical_live} use the Qwen3.8-27B planner. We use the same Qwen3.8-27B planner for our \capbench runs and for both skill ablations (Appendix~\S\ref{sec:appendix:emergence} and \S\ref{sec:appendix:skill_library}). For our Gemini programs we prompt for a yaml behavior tree when generating the BT and a list of phrases when generating prompts. Apart from setting the temperature and thinking budget they use the SDK defaults for all output limits. The system prompt when generating BTs describes the primitive grammar as well as the YAML output format. When generating prompts we ask it for 3 color-and-shape prompts for each object and receive back a json list of strings.

\subsection{Depth Sources}
\label{sec:appendix:da3}

For simulation we read ground-truth depth from the renderer. For our physical experiments we use time of flight depth from two Azure Kinect DKs (sideview master and birdview subordinate) which we read using \texttt{pyk4a}. Our wrist mounted RealSense D435I additionally gives us stereo depth which we use for close-range grasp refinement.

\subsection{Pyroki IK Settings}
\label{sec:appendix:pyroki}

For our Table~\ref{tab:libero_main} experiments we disabled Pyroki and instead fell back to the executor's built in damped least-squares 6-DOF IK for moves that require an orientation constraint (such as pulling a drawer open or grasping the rim of a plate). For \capbench we enable PyRoki~\citep{kim2025pyroki}, a JAX-based toolkit for robot kinematic optimization, and use its IK for joint-space descent. We found the \texttt{panda\_hand\_tcp} link offset by $6.9$~mm in the z direction from the \texttt{grip\_site} used by LIBERO, and correct for this when solving. We use the default weights from Pyroki of $w_{\text{pos}}{=}1.0$ and $w_{\text{ori}}{=}0.5$. We enable both joint limit and self collision constraints on the Franka chain. If our constraints are empty we use MuJoCo's Jacobian pseudo-inverse for a faster solve.

\subsection{OSC Controller}
\label{sec:appendix:osc}

We use the Robosuite~\citep{zhu2020robosuite} operational-space controller \texttt{OSC\_POSE} with fixed impedance and a 6-D pose action. We set the default stiffness to $k_p = 150$ across all six axes with critical damping. Movements of $5$~cm take less than $0.5$~s to settle. When falling back to joint space we use Robosuite's \texttt{JOINT\_POSITION} controller with max delta of $0.05$~rad per step. The controller layer enforces this cap instead of the LLM. On our real robot pipeline we also enforce a max TCP delta of $0.02$~m per step.

\subsection{BT Executor}
\label{sec:appendix:exec}
At the end of an episode we call \texttt{env.check\_success()} once to determine if it was successful. For the \rats-protocol row of Table~\ref{tab:libero_main}, the three-task study of Table~\ref{tab:libero_dyn}, as well as the perception ablation, \capbench runs, and both skill ablations we poll the success predicate after each attempt during eval as described by \rats, where the success check for the task after each attempt determines whether we retry the task. The strict row of Table~\ref{tab:libero_main} and the displacement studies of Tables~\ref{tab:dyn_main} and~\ref{tab:dyn_matched} read it only once, after the episode ends. In the strict row an attempt fails when the plan's own checks fail. In a plan without checks it fails when the last grasp is empty or slips. Both rows of Table~\ref{tab:libero_main} keep every other read of the predicate off during the episode. In the \rats-protocol row only the check after each attempt is read. Each primitive returns a bool through a post condition that determines if it succeeded (e.g. grasp checks the gap between fingers is below a contact threshold). If this returns false during the three-task study of Table~\ref{tab:libero_dyn} we call our recovery layer allowing for up to 2 retries before failing closed. Before the gripper closes, the executor checks that the hand reached the grasp pose, and a blocked descent ends the approach. A placement check fails when nothing was held at release or the jaws did not open. Otherwise it re-detects the object and tests whether it lies over the target, and it passes when the object is not seen. During our Table~\ref{tab:libero_main} simulation runs, grasp and placement checks written into the plan act through its retry and fallback nodes, and the robot recovers between attempts as described in \S\ref{sec:fallback}. The executor ignores LIBERO's \texttt{done} flag during the episode, so every motion and release runs to completion. On our UR10e executor each primitive has the following timeouts: \texttt{move\_to\_keypoint} $4$~s, \texttt{move\_relative} $2$~s, \texttt{grasp}/\texttt{release} $1$~s, \texttt{insert} $6$~s, \texttt{push\_object} $4$~s, \texttt{open\_drawer} $5$~s and \texttt{wipe} $10$~s. On timeout we call our recovery layer and continue the trial. Our simulation executors also enforce step budgets for each primitive. When executing a plan on our UR10e executor we concatenate consecutive Cartesian waypoints into one Ruckig time-optimal trajectory (\emph{legato} execution) instead of going to each point separately.

\section{LIBERO-PRO Per-Task Results (Ablation Configuration)}
These are the per-task vectors of the configuration used for the perception ablation. The per-task vectors for Table~\ref{tab:libero_main} are in Appendix~\S\ref{sec:appendix:provenance}.
\label{sec:appendix:per-task}

\begin{figure}[h]
\centering
\includegraphics[width=\linewidth]{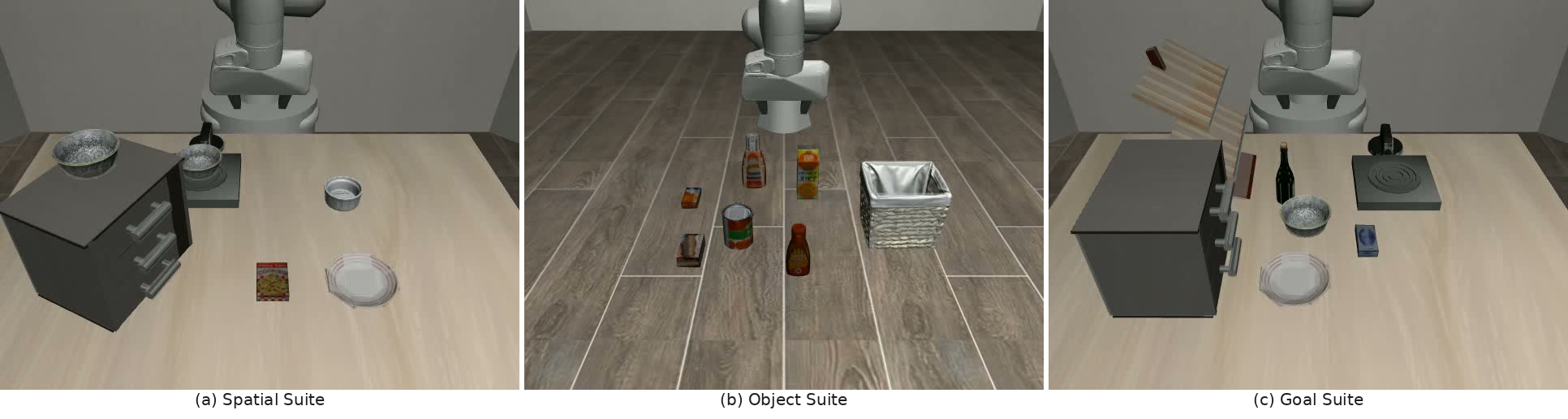}
\caption{Representative \libpro simulation environments. (a) Spatial suite: bowl pick-and-place with position perturbation across kitchen fixtures. (b) Object suite: grocery items into a basket with object substitution. (c) Goal suite: multi-step tasks involving drawers, stove, and cabinet with goal redefinition.}
\label{fig:libero_envs}
\end{figure}

\begin{figure}[h]
\centering
\includegraphics[width=0.78\linewidth]{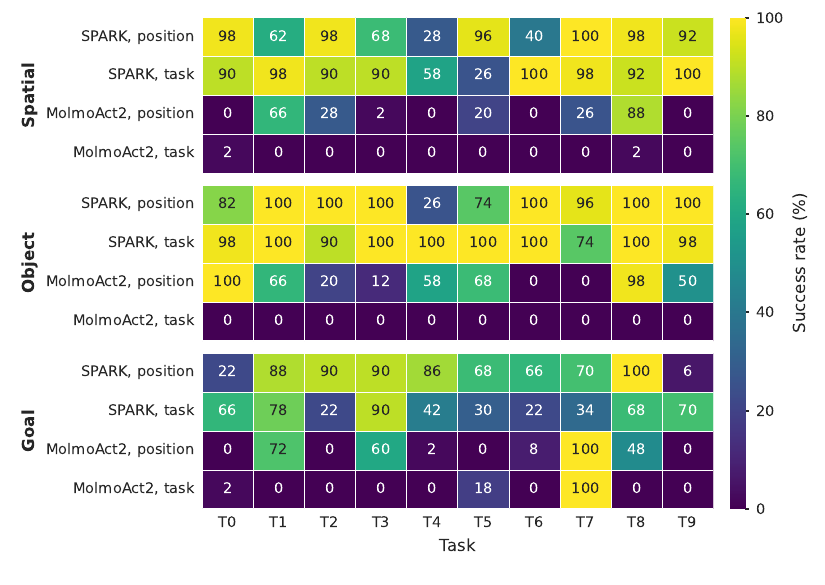}
\caption{Per-task heatmap of position and task success rates ($\%$) across three \libpro suites. The \spark rows use the \rats protocol of Table~\ref{tab:libero_main}. Tasks T0 to T9 follow the row order of Tables~\ref{tab:appendix:object_v5}--\ref{tab:appendix:goal_v5}. \spark outperforms MolmoAct2 on task-level perturbation across all three suites. MolmoAct2 achieves higher position accuracy on specific object-suite tasks but scores near zero on task perturbation.}
\label{fig:heatmap}
\end{figure}

Figure~\ref{fig:libero_envs} shows representative scenes from each \libpro suite.
Tables~\ref{tab:appendix:libero-spatial}--\ref{tab:appendix:libero-goal} give per-task \spark (adaptive) success rates across the spatial, object, and goal suites of \libpro ($50$ trials per task), and Figure~\ref{fig:heatmap} compares the per-task results of the Table~\ref{tab:libero_main} run with MolmoAct2.

Table~\ref{tab:appendix:molmo-spatial} gives the corresponding MolmoAct2-LIBERO rates from our evaluation, as no published \libpro results exist for MolmoAct2.

\begin{table}[h]
\centering\small
\setlength{\tabcolsep}{4pt}
\begin{minipage}[t]{0.48\linewidth}
\centering
\caption{\spark adaptive: spatial suite per-task success rates ($\%$, $50$ trials).}
\label{tab:appendix:libero-spatial}
\begin{tabular}{lcc}
\toprule
Task & Pos & Task \\
\midrule
between plate \& ramekin   & $62$ & $98$ \\
next to ramekin            & $96$ & $72$ \\
from table center          & $82$ & $40$ \\
on cookie box              & $88$ & $74$ \\
in top drawer              & $0$  & $70$ \\
on ramekin                 & $90$ & $66$ \\
next to cookie box         & $4$ & $82$ \\
on stove                   & $32$ & $56$ \\
next to plate              & $16$ & $90$ \\
on wooden cabinet          & $66$ & $76$ \\
\midrule
Mean & $53.6$ & $72.4$ \\
\bottomrule
\end{tabular}
\end{minipage}\hfill
\begin{minipage}[t]{0.48\linewidth}
\centering
\caption{\spark adaptive: object suite per-task success rates ($\%$, $50$ trials).}
\label{tab:appendix:libero-object}
\begin{tabular}{lcc}
\toprule
Task & Pos & Task \\
\midrule
alphabet soup       & $22$ & $30$  \\
cream cheese        & $28$ & $40$  \\
salad dressing      & $20$ & $88$  \\
bbq sauce           & $76$ & $18$  \\
ketchup             & $6$  & $50$  \\
tomato sauce        & $76$ & $92$  \\
butter              & $20$ & $8$   \\
milk                & $86$ & $6$   \\
chocolate pudding   & $0$ & $32$  \\
orange juice        & $64$ & $0$   \\
\midrule
Mean & $39.8$ & $36.4$ \\
\bottomrule
\end{tabular}
\end{minipage}

\vspace{8pt}
\begin{minipage}[t]{0.42\linewidth}
\centering
\caption{\spark adaptive: goal suite per-task success rates ($\%$, $50$ trials).}
\label{tab:appendix:libero-goal}
\begin{tabular}{lcc}
\toprule
Task & Pos & Task \\
\midrule
open middle drawer          & $0$   & $0$  \\
bowl on stove               & $100$ & $0$  \\
wine bottle on cabinet      & $42$  & $12$ \\
bowl in top drawer          & $0$   & $0$  \\
bowl on cabinet             & $82$  & $0$  \\
push plate to stove         & $0$   & $40$ \\
cream cheese in bowl        & $56$  & $14$ \\
turn on stove               & $10$  & $2$  \\
bowl on plate               & $98$  & $14$ \\
wine bottle on rack         & $12$  & $58$ \\
\midrule
Mean & $40.0$ & $14.0$ \\
\bottomrule
\end{tabular}
\end{minipage}\hfill
\begin{minipage}[t]{0.56\linewidth}
\centering
\caption{MolmoAct2-LIBERO per-task success rates ($\%$, $50$ trials). These are from our evaluation using the public MolmoAct2 checkpoint.}
\label{tab:appendix:molmo-spatial}
\begin{tabular}{lcccccc}
\toprule
& \multicolumn{2}{c}{Spatial} & \multicolumn{2}{c}{Object} & \multicolumn{2}{c}{Goal} \\
\cmidrule(lr){2-3}\cmidrule(lr){4-5}\cmidrule(lr){6-7}
Task & Pos & Task & Pos & Task & Pos & Task \\
\midrule
T0 & $0$   & $2$   & $100$ & $0$ & $0$   & $2$   \\
T1 & $66$  & $0$   & $66$  & $0$ & $72$  & $0$   \\
T2 & $28$  & $0$   & $20$  & $0$ & $0$   & $0$   \\
T3 & $2$   & $0$   & $12$  & $0$ & $60$  & $0$   \\
T4 & $0$   & $0$   & $58$  & $0$ & $2$   & $0$   \\
T5 & $20$  & $0$   & $68$  & $0$ & $0$   & $18$  \\
T6 & $0$   & $0$   & $0$   & $0$ & $8$   & $0$   \\
T7 & $26$  & $0$   & $0$   & $0$ & $100$ & $100$ \\
T8 & $88$  & $2$   & $98$  & $0$ & $48$  & $0$   \\
T9 & $0$   & $0$   & $50$  & $0$ & $0$   & $0$   \\
\midrule
Mean & $23.0$ & $0.4$ & $47.2$ & $0.0$ & $29.0$ & $12.0$ \\
\bottomrule
\end{tabular}
\end{minipage}
\end{table}

\section{CaP-Bench Per-Task Details}
\label{sec:appendix:capbench}

We report the BT utilized for each task in \capbench, the number of trials, their success criterion, and the dominant failure mode observed. All \spark and \capx results in this section use $100$ trials per task. \rats numbers are taken from its $50$-trial RoboSuite skill-transfer evaluation. The \spark column uses the Qwen3.8-27B planner, robosuite 1.5.2 and the Pyroki joint-space descent, as does the skill-emergence ablation (Appendix~\S\ref{sec:appendix:emergence}).

\begin{table}[h]
\centering
\caption{\capbench results ($\%$). \capx numbers are from \citet{fu2026capx} (its full $3$-model ensemble), and \rats numbers are from \citet{rats2026playful} (its ``CaP-Agent0 $+$ \textsc{RATs} skills'' transfer column, $50$ trials/task). NutAssemblySquare is \capx's Peg Insert.}
\label{tab:capbench}
\begin{tabular}{lccc}
\toprule
Task & \capx (3M) & \rats & \spark \\
\midrule
Lift            & $97$ & $84$ & $\mathbf{100}$ \\
Stack           & $\mathbf{98}$  & $60$ & $93$ \\
CubeRestack     & $89$  & $46$ & $\mathbf{97}$ \\
Wipe            & $\mathbf{100}$  & $\mathbf{100}$ & $57$ \\
NutAssemblySquare & $0$       & $0$ & $\mathbf{2}$ \\
TwoArmLift      & $74$  & $34$ & $\mathbf{77}$ \\
TwoArmHandover  & $20$  & $20$ & $\mathbf{30}$ \\
\bottomrule
\end{tabular}
\end{table}

Lift ($100\%$). There is only one short BT plan for this task: \texttt{move\_to\_keypoint(cube, offset\_z=0.05)} $\to$ \texttt{grasp} $\to$ \texttt{move\_relative(dz=0.15)}. The task is considered complete if the cube is $4$\,cm off of the table.

Stack ($93\%$). \texttt{move\_to\_keypoint(cube\_A)} $\to$ \texttt{grasp} $\to$ \texttt{move\_to\_keypoint(cube\_B, offset\_z=0.08)} $\to$ \texttt{release}. We consider the task complete if cube A is stacked on cube B with the contact normal facing upward. All $7$ failures were due to a failed grasp resulting in no object being placed.

CubeRestack ($97\%$). Uses the Stack plan but since cube A is on top of a third cube we just check that cube A is on top of cube B regardless of what it was on. Failed due to a failed grasp $3$ times.

Wipe ($57\%$). The planner always selects \texttt{constrained\_scrub} for all plans which will scrub back and forth using a normal force over the dirt found using SAM3 (some plans scrub over $2$ regions). Out of the $43$ failures, $24$ decided to stop scrubbing because they did not detect the dirt patch getting smaller while there are still markers on the table. $9$ exhausted the number of steps the scrub could take and $9$ reached the early end condition of robosuite's Wipe task. The last failure had another reason for not completing.

NutAssemblySquare ($2\%$). None of our baseline methods can solve this task. The planner creates a pick and place sequence that ends in an \texttt{insert} action. SAM3 detects the nut as a nut in $90/100$ rollouts and the square peg in $98/100$ rollouts. Since the nut has a hollow center our grasp action grasps towards the side of the nut rather than the center of mass of the mask (which is on the hole in the middle). \spark is able to grasp the nut in $55/100$ rollouts and bring it over the square peg in $53/100$ rollouts. However it only successfully inserts the nut onto the square peg $2/100$ times. To insert the nut you have to line it up with the peg and maintain your pose while making contact with the peg. The skills we have available to the planner cannot do this consistently. If we write a custom NutAssemblySquare routine that utilizes SAM3 detections and some shape knowledge and full-pose joint space control instead of the benchmark's OSC controller we are able to insert the nut $41/50$ times ($82\%$). This is not included in our results for \spark.

TwoArmLift ($77\%$ over $100$ rollouts) and TwoArmHandover ($30\%$) are both created by our planner. TwoArmLift contains a parallel node with $2$ \texttt{pick\_with\_arm} actions. One for each handle (left and right). These actions contain a width check on the closed gripper that confirms it is holding the handle. Once both arms are gripping the handles they sync at a \texttt{sync\_barrier} and lift the pot together. Each arm will grasp from the top down and have its wrist yawed to be on the small handle bar. Each arm will try to re-grasp its respective handle while the other arm holds. If either arm fails to grasp at the barrier the lift is aborted and the pot remains on the table. We use adaptive perception which will combine similar detections into one. Since this task requires us to detect the left and right handle we disable this feature and prompt for each handle separately. $19/23$ failed attempts were due to the lift being aborted, meaning that both arms did not have the pot after their re-grasp attempts, most likely due to a bad centroid from the small handle bar using SAM3. The other $4$ times the pot slipped off the arms while lifting.

TwoArmHandover ($30\%$). This plan uses a \texttt{handoff} primitive, where arm A will grasp the hammer and move it to a meeting point. Then arm B will grasp the hammer and arm A will release. Arm A will not release the hammer until arm B has grasped it so at all times the hammer is held by one arm. Similar to TwoArmLift arm A will grasp with a width check and will try to re-grasp. After failing to grasp it will also sweep around its wrist azimuth and descend at different depths. $35$ times arm B was not on the hammer when arm A released it. This could be due to arm B having a bad grasp on the vertical handle coming across the table. $12$ times arm A failed to grasp and $10$ times it dropped the hammer. $9$ times arm B could not find a grasp and $4$ times arm A was still on the hammer when the rollout ended.

\spark performs lower on Wipe ($57\%$ vs.\ $100$) because the only way to tell if you have wiped the spill is by looking at the scene after your action. \capx compares the images at each turn and knows if it has wiped the spill. Our plans are predetermined before we know if we have wiped the spill or not, and the scrub will stop once the detected dirt patch stops getting smaller. The two arm tasks are difficult for all methods. \spark is able to create these plans using primitives and achieve $77\%$ (\capx achieves $74$ and \rats $34$) on TwoArmLift and $30\%$ (both \capx and \rats achieve $20$) on TwoArmHandover.

\section{BT Grammar Reference}
\label{sec:appendix:grammar}

\begin{figure}[t]
\centering
\includegraphics[width=\linewidth]{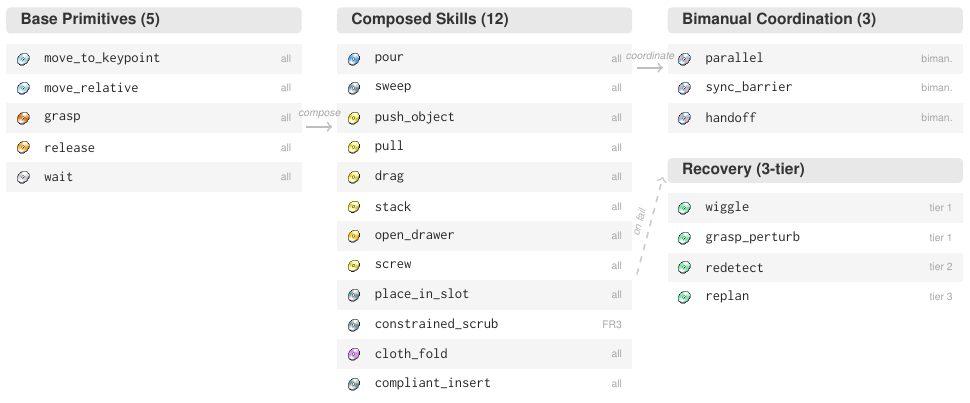}
\caption{\spark primitive library. We show the $5$ base primitives (left) and how they combine to form manipulation and tool use skills (middle, a representative subset of the more than $30$ registered skills) as well as bimanual nodes (right). There is a hierarchical system for recovering from failures, starting from perturbations in place (\texttt{wiggle}, \texttt{grasp\_perturb}) to re-grounding the perception (\texttt{redetect}) to regenerating the plan (\texttt{replan}). Only the first $2$ tiers will occur within one attempt at executing. In Table~\ref{tab:libero_main} only the re-plan between attempts occurs (\S\ref{sec:fallback}).}
\label{fig:primitive_menu}
\end{figure}

Figure~\ref{fig:primitive_menu} gives an overview of the primitive library. Below is a sample of the YAML BT DSL in BNF-like form. A program is a \texttt{task} string and a \texttt{tree} whose root is a \texttt{sequence} (or \texttt{parallel}) node whose children are typed primitive invocations. For planner output parsed (as YAML) in Table~\ref{tab:libero_main}, malformed extension fields are dropped and every object label is snapped to a detected label or requested again once. Node types are not validated, and the simulation executor skips unknown types. It runs a \texttt{retry} node for at most $3$ attempts and a \texttt{fallback} node for at most $3$ branches, with at most $6$ re-attempts in one plan. A plan that fails to parse is counted as failing the trial. The real-robot executor runs selector and retry nodes to completion. For displacement study trials, a false return condition triggers a retract-and-re-render cycle that re-grounds perception without re-querying the LLM. For trials in Table~\ref{tab:libero_main}, every recovery will consist of $1$ new LLM call and executing the plan. The grammar will remain the same throughout recoveries.

\noindent
\begin{minipage}{\linewidth}
\small
\begin{verbatim}
<program>      ::= "task:" <text> NL "tree:" NL <node>
<node>         ::= "type:" <ctrl> NL "children:" NL <child>+
<ctrl>         ::= "sequence" | "parallel"
<child>        ::= "- type:" <prim_id> NL "  params:" NL <arg>*
                 | "- " <node>
<prim_id>      ::= "move_to_keypoint" | "move_relative" | "grasp"
                 | "release" | "wait" | "grasp_se3" | "insert"
                 | "push_object" | "open_drawer" | "pour"
                 | "sweep" | "wipe" | "constrained_scrub"
                 | "pull" | "drag" | "stack" | "screw"
                 | "pick_with_arm" | "sync_barrier" | "handoff" | ...
<arg>          ::= "    " <slot> ": " <value> NL
\end{verbatim}
\end{minipage}

\vspace{0.5em}
Per-primitive slot signatures and defaults are listed in Table~\ref{tab:appendix:primitives}. Each skill is a Python function that is registered with its typed arguments and its post condition. The executor calls the function at the matching node of the tree. To add a skill, we register a new function, and the planner can then use it with no training.

\begin{table}[h]
\centering
\caption{Typed slots for each \spark primitive. Required slots are marked with ($\star$). Five base primitives (\texttt{move\_to\_keypoint}, \texttt{move\_relative}, \texttt{grasp}, \texttt{release}, \texttt{wait}) compose into the manipulation and tool-use skills below. The table lists these five and a representative subset of the more than thirty registered skills. The scalar slot values (offsets, forces, angles, durations such as \texttt{wait}'s) are emitted by the planner model (temperature $0$ in simulation, $0.3$ on hardware): the keypoint label fixes only the spatial anchor, so the planner sets the scalar values directly.}
\label{tab:appendix:primitives}
\small
\setlength{\tabcolsep}{2pt}
\begin{tabular}{lll}
\toprule
Primitive & Slots & Default \\
\midrule
\texttt{move\_to\_keypoint} & \texttt{keypoint\_label}$^\star$, \texttt{offset\_x/y/z} & all $=0$ \\
\texttt{move\_relative}     & \texttt{dx}, \texttt{dy}, \texttt{dz}       & all $=0$ \\
\texttt{grasp}              & \texttt{force}                              & \texttt{force}$=100$ \\
\texttt{release}            & --                                          & -- \\
\texttt{wait}               & \texttt{duration}                           & \texttt{duration}$=0.5$ \\
\texttt{grasp\_se3}         & \texttt{keypoint\_label}$^\star$, \texttt{strategy}, \texttt{force} & \texttt{force}$=60$ \\
\texttt{insert}             & \texttt{keypoint\_label}$^\star$            & -- \\
\texttt{push\_object}       & \texttt{keypoint\_label}$^\star$, \texttt{push\_direction}, \texttt{push\_distance} & $0.10$ \\
\texttt{open\_drawer}       & \texttt{keypoint\_label}$^\star$, \texttt{pull\_distance} & $0.15$ \\
\texttt{pour}               & \texttt{target\_label}$^\star$, \texttt{pour\_angle\_rad} & $1.57$ \\
\texttt{sweep}              & \texttt{target\_label}$^\star$, \texttt{object\_labels}, \texttt{n\_passes} & \texttt{n\_passes}$=2$ \\
\texttt{wipe}               & \texttt{keypoint\_label}$^\star$            & -- \\
\texttt{constrained\_scrub} & \texttt{target\_label}$^\star$, \texttt{force\_n}, \texttt{n\_cycles}, \texttt{pattern} & \texttt{force\_n}$=8$ \\
\texttt{pull}               & \texttt{keypoint\_label}$^\star$, \texttt{pull\_distance}   & $0.15$ \\
\texttt{drag}               & \texttt{keypoint\_label}$^\star$, \texttt{target\_label}$^\star$ & -- \\
\texttt{stack}              & \texttt{target\_label}$^\star$, \texttt{force\_threshold}, \texttt{max\_descent} & -- \\
\texttt{screw}              & \texttt{angle}, \texttt{linear\_distance}      & \texttt{angle}$=\pi$ \\
\bottomrule
\end{tabular}
\end{table}

\section{Skill-Emergence Ablation}
\label{sec:appendix:emergence}

Our approach claims that we can positionally control a robot with $5$ primitives (\texttt{move\_to\_keypoint}, \texttt{move\_relative}, \texttt{grasp}, \texttt{release}, \texttt{wait}) and that we can compose these primitives into skills with our grammar. We ablate this by checking that we are able to complete tasks without writing any code for the task. We achieve this with $2$ hooks that we can intervene with reproducibly. 1) We have an environment hook that will remove a skill from the grammar that we show the planner (meaning the BT planner cannot access this skill). 2) We will prompt the planner to use the base primitives to complete the motion they desire in context before making the BT. We then test our \capbench tasks that require this skill and see if we can complete them with base primitives (Table~\ref{tab:emergence}). From the executor logs we can verify that we are not calling any nodes from the skill when running on base primitives.

\begin{table}[h]
\centering
\caption{Skill-emergence ablation. When removing the composed skill from the grammar the planner must use the $5$ base primitives to complete the task. It is able to create the pick and place tasks successfully. The Wipe task gets a valid plan that does not meet the force-and-coverage criteria for the task.}
\label{tab:emergence}
\begin{tabular}{lccc}
\toprule
Task & Composed skill withheld & Base-only success (\%) & Trials \\
\midrule
Lift  & \emph{(none; already base)} & $100$ & $10$ \\
Stack & \texttt{stack}              & $100$ & $10$ \\
Wipe  & \texttt{constrained\_scrub} & $0$   & $20$ \\
\bottomrule
\end{tabular}
\end{table}

We can see that the planner is able to compose a skill that can be completed with our base primitives for our kinematic pick and place tasks. Since we do not have an executor for \texttt{stack} on our platform we also use base primitive plans for the \capbench Stack result in Table~\ref{tab:capbench}. We can see that the plan created when withholding \texttt{stack} is \texttt{move\_to\_keypoint}, \texttt{grasp}, \texttt{move\_relative}, \texttt{move\_to\_keypoint} (place offset), \texttt{release}.

Wipe is where we hit the limit. Using only base primitives we are able to generate a valid BT. The planner understands that it should create a boustrophedon pattern of \texttt{move\_relative} to wipe the dirt off the surface. But we score $0\%$ on this task where the composed skill \texttt{constrained\_scrub} scores $57\%$ (Appendix~\ref{sec:appendix:capbench}). The planner can't write closed loop control from just language: descending until contact with a force controller, knowing how far to move in a raster pattern based on the size of the dirty area, and when to re-wipe. If you need force control and/or control over coverage you need to use one of our composed skills. This is why we have over $30$ different skills with types.

\section{Skill-Library Ablation}
\label{sec:appendix:skill_library}

This ablation measures how much of the \libpro result comes from the registered skills. We run four skill libraries with the planner, executor and perception of Table~\ref{tab:libero_main}, under the \rats protocol, on all six settings with initial states $0$ to $9$ ($600$ trials per library, Table~\ref{tab:skill_library}). \emph{Full} has all registered skills, and its row is the matching $600$ trials of the \rats-protocol run of Table~\ref{tab:libero_main}. \emph{Generic skills only} has the 5 base primitives, \texttt{grasp\_se3} and the two checks (\texttt{verify\_grasp}, \texttt{verify\_placed}). \emph{5 base primitives} has only \texttt{move\_to\_keypoint}, \texttt{move\_relative}, \texttt{grasp}, \texttt{release} and \texttt{wait}. These two libraries give the same motions on \libpro and differ in the prompt text. We designed \texttt{open\_drawer}, \texttt{push\_object} and the knob turn (\texttt{screw}, \texttt{turn\_knob}) on LIBERO scenes. \texttt{move\_to\_keypoint} and \texttt{grasp} also have two branches for wide objects, a rim pinch for bowls and one for plates. The last row removes these two branches and selects the grasp from the mask of the object: the gripper pinches the rim when the mask is wider than the open gripper. This rule uses no object names.

\begin{table}[h]
\centering
\caption{Skill-library ablation on \libpro, \rats protocol. Successes of $200$ per suite (position and task settings, $10$ tasks, initial states $0$ to $9$) and of $600$ in total. The last column is the paired difference from the full library in percentage points, with a 95\% interval from resampling the $60$ setting-task pairs.}
\label{tab:skill_library}
\small
\setlength{\tabcolsep}{4pt}
\begin{tabular}{lccccc}
\toprule
Library & Object & Goal & Spatial & All & vs.\ full \\
\midrule
Full & $182$ & $127$ & $155$ & $464$ & \\
Generic skills only & $181$ & $80$ & $160$ & $421$ & $-7.2$ $[-13.8, -1.5]$ \\
5 base primitives & $182$ & $85$ & $157$ & $424$ & $-6.7$ $[-13.5, -1.0]$ \\
Base, grasp chosen from mask width & $185$ & $81$ & $161$ & $427$ & $-6.2$ $[-13.2, -0.2]$ \\
\bottomrule
\end{tabular}
\end{table}

The 5 base primitives are sufficient for pick and place: $423$ of the $520$ pick-and-place trials succeed with them, and $415$ succeed with the full library. The loss is in the goal suite (Table~\ref{tab:skill_library}). In the runs of Table~\ref{tab:libero_main}, $788$ of the $6{,}000$ first plans use a named skill, and all of them are in the goal suite: \texttt{open\_drawer} ($400$), \texttt{push\_object} ($200$), \texttt{screw} ($171$), \texttt{tilt\_wrist} ($11$) and \texttt{turn\_knob} ($6$).

\section{Failure-Mode Analysis}
\label{sec:appendix:failures}

\paragraph{Spatial-Pos.} The biggest failure mode is the gripper occluding objects in the wrist camera. The object will be within the occlusion zone of the closed gripper, causing adaptive's wrist-mask check to reject a valid detection. Since recoveries will re-render the same prompts we will continue to not detect the object.

\paragraph{Spatial-Task.} Same failure modes as Spatial-Pos. The spatial rewrite can also point to a second object that looks the same, and the prompt may detect both objects (\eg ``black bowl'' detects the target and the other bowl).

\paragraph{Object-Pos.} Objects that look similar can be merged by our $30$\,px centroid threshold. Not leaving enough room for pre-grasp on tall objects causes our gripper to collide on the way down. The place pose underestimates the rim height of the container.

\paragraph{Object-Task.} Adaptive will generate $3$ variants for an object that has been renamed. These can split one object up between several detections from SAM3. Sometimes our call to generate variants will generate prompts that don't describe the object correctly.

\paragraph{Goal-Pos.} The $z$ value of our destination keypoint can be off if the surface is see-through or shiny (glass plate, polished stove). The BT will be correct but \texttt{insert} times out while trying to line up.

\paragraph{Goal-Task.} Adaptive generates noisier prompts than our fair baseline for rewritten tasks.

\begin{figure}[t]
\centering
\includegraphics[width=0.8\linewidth]{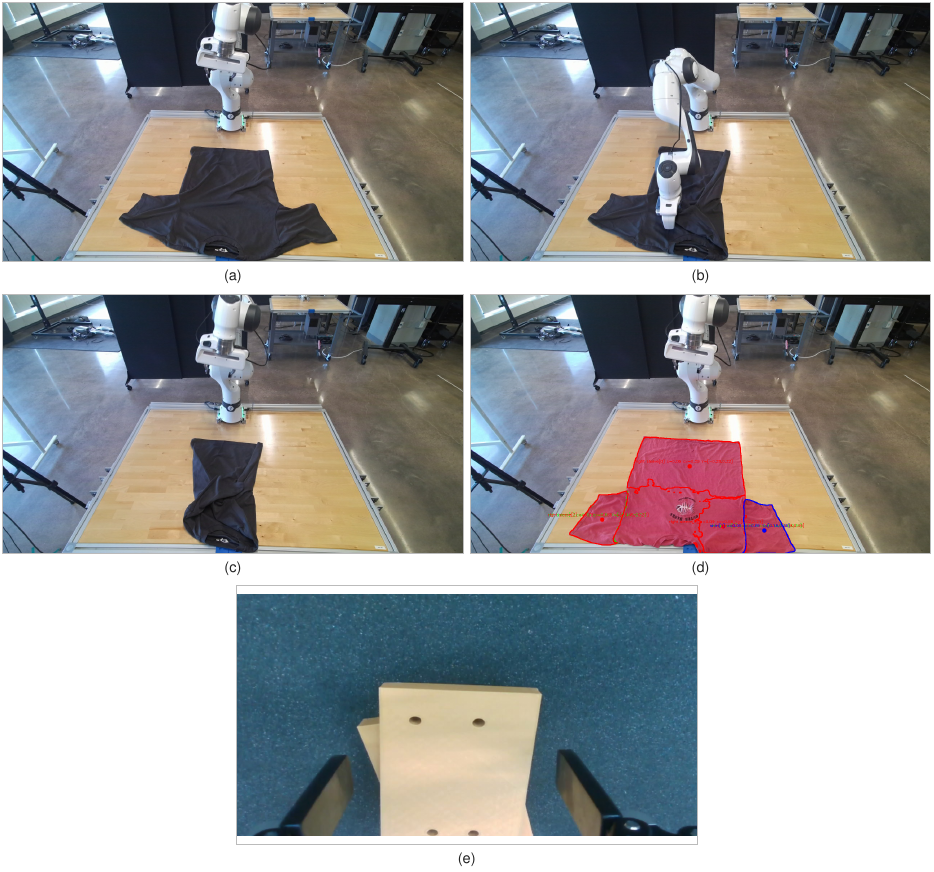}\\[4pt]
\includegraphics[width=0.42\linewidth]{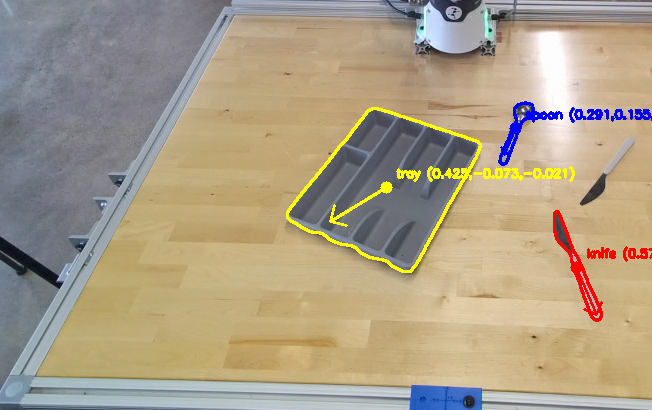}
\caption{Examples of perception failures. (a)--(d) Folding a garment: (a) a dark shirt flat on the surface, (b) sleeves folded correctly, (c) the hem line of the dark shirt is not detected, (d) the sleeve masks on the red shirt extend over the middle and carry both sleeve labels. (e) Wrist-camera refinement offsets the grasp on a large gold block. (f) The tray major axis from the OBB is misaligned, which gives poor slot placement.}
\label{fig:failures}
\end{figure}

\section{Physical Experiments: Per-Trial Logs}

\label{sec:appendix:physical}
Two complete behavior trees from these platforms are in Appendix~\S\ref{sec:appendix:bt-examples}.
Table~\ref{tab:physical} lists the eleven cells, and Table~\ref{tab:physical_live} gives the UR10e trials with live displacement and their timing.

\begin{table}[h]
\centering
\caption{Physical success rates ($\%$, $20$ trials per task, objects randomly placed and rotated per trial). The Planner column names the model that wrote the plans of each row.}
\label{tab:physical}
\small
\setlength{\tabcolsep}{3pt}
\begin{tabular}{lllc}
\toprule
Embodiment & Task & Planner & Success ($\%$) \\
\midrule
UR10e & Utensils in bowl       & Qwen3.8-27B (4-bit) & $90$ \\
UR10e & Utensils in tray       & Qwen3.8-27B (4-bit) & $\mathbf{100}$ \\
UR10e & Plushie in bowl        & Qwen3.8-27B (4-bit) & $\mathbf{100}$ \\
UR10e & Stack blocks           & Qwen3.8-27B (4-bit) & $95$ \\
\midrule
FR3   & Utensils in tray       & Gemini 3.5 Flash & $80$ \\
FR3   & Sponge-wash plate      & Gemini 3.5 Flash & $\mathbf{100}$ \\
FR3   & Mug pour               & Gemini 3.5 Flash & $65$ \\
FR3   & Sweep to dustpan       & Gemini 3.5 Flash & $55$ \\
FR3   & T-shirt fold           & Gemini 3.5 Flash & $50$ \\
\midrule
Bimanual & T-shirt fold        & Gemini 3.5 Flash & $30$ \\
Bimanual & Silverware sort     & Gemini 3.5 Flash & $60$ \\
\midrule
\multicolumn{3}{l}{Mean (11 cells)} & $75$ \\
\bottomrule
\end{tabular}
\end{table}

\begin{figure}[h]
\centering
\includegraphics[width=0.9\linewidth]{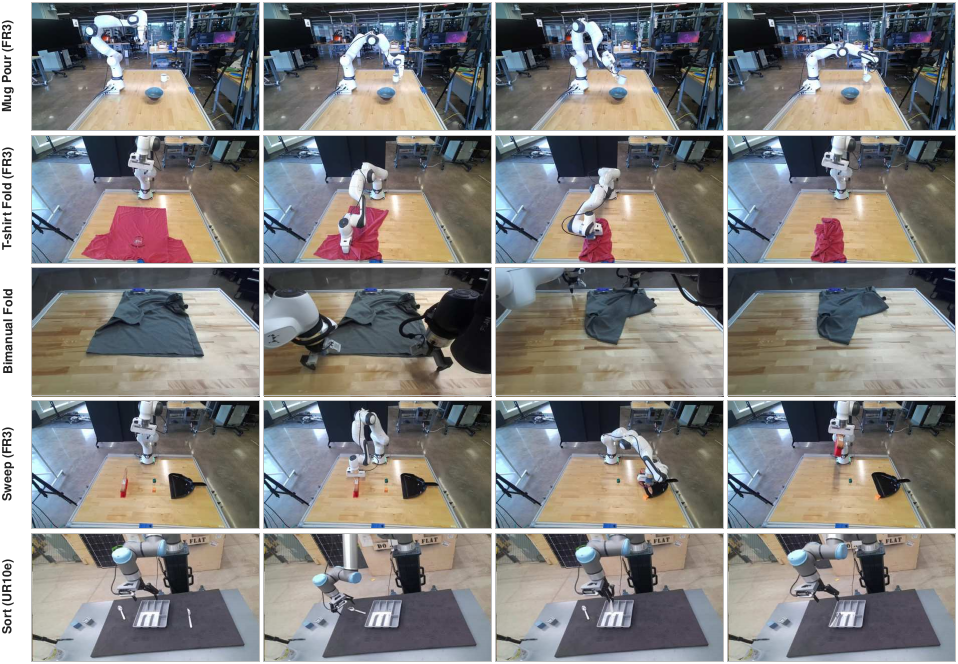}
\caption{Physical execution sequences. Shown here are $4$ frames from each task being executed on our physical platforms: pouring from a mug (FR3), folding a t-shirt (FR3), folding bimanually (Panda + FR3), sweeping into a dustpan (FR3) and sorting silverware (UR10e).}
\label{fig:qualitative}
\end{figure}

There were $9$ tasks spanning $3$ embodiments ($11$ total task-embodiment combinations). We randomize task object placement for each trial and have some tasks switch between object types between runs (\ie different plushies or sets of utensils). All runs of Table~\ref{tab:physical} use the same BT language and perception stack (SAM3). The UR10e rows use the Qwen3.8-27B planner of Table~\ref{tab:libero_main} on a separate GPU, and the FR3 and bimanual rows use Gemini 3.5 Flash.

\subsection*{UR10e + Robotiq 2F-85}

\textbf{Utensils in bowl} ($90\%$, $18/20$). Each trial places one spoon. In one failure the spoon ended beside the bowl after a retry. In the other the spoon ended behind the bowl, where only one camera could see it, and the planner did not move the bowl out of the way. We reset the scene between each trial manually. The executor itself can invert this placement, grasping each utensil out of the bowl and releasing it onto the table, which is an ENPIRE-style automatic reset~\citep{xiao2026enpire} and lets collection run unattended.

\textbf{Utensils in tray} ($100\%$, $20/20$). Counts successes only where the robot yaw-aligned the utensil with the tray direction and placed each type of utensil in its own slot. Utensils of the same type share a slot. A trial places one to three utensils.

\textbf{Plushie in bowl} ($100\%$, $20/20$). Saturated on different plushie types. The top down grasp primitive works on the soft plushies.

\textbf{Stack blocks} ($95\%$, $19/20$). In the one failure the gripper made a hard contact on the descent and the arm stopped responding.

\subsection*{Franka FR3}

\textbf{Utensils in tray} ($80\%$, $16/20$). Main failure comes from SAM3 misclassifying the spoon handle as a knife and placing it in the wrong tray. Got a $100\%$ success rate on UR10e with the Qwen planner. The wider Robotiq jaw may be able to accommodate the shape of the utensils better.

\textbf{Sponge-wash plate} ($100\%$, $20/20$). Was able to consistently use our composition of skills (\texttt{grasp\_se3} $\to$ \texttt{tool\_grip\_pose} $\to$ \texttt{constrained\_scrub}) within the FR3 work space.

\textbf{Mug pour} ($65\%$, $13/20$). Failure occurs when the depth sensor struggles with the thin portion of the mug handle causing an inaccurate guess of the handle position, and the gripper misses the handle. Another failure mode is Gemini choosing to use \texttt{tilt\_wrist} (roll) instead of \texttt{pour} (pitch) which causes the mug to spill out to the side.

\textbf{Sweep to dustpan} ($55\%$, $11/20$). Failures occur from grabbing the brush at the wrong position (too high on the bristles) and not sweeping towards the dustpan at a steady angle. If we had another arm that could hold the dustpan at an angle this task should be more successful.

\textbf{T-shirt fold} ($50\%$, $10/20$). We couldn't get a single successful run with black or grey shirts: SAM3 returns inconsistent hem labels on dark fabrics. Sleeve folding was still successful. Coordinate filtering could alleviate this issue. It successfully folds light-colored shirts if sleeve detection succeeds.

\subsection*{Bimanual Franka (Panda + FR3)}

\textbf{T-shirt fold} ($30\%$, $6/20$). Folding the sleeves succeeds in $75\%$ ($15/20$) of trials. Folding all $3$ parts ($2$ sleeves and bottom hem) fails because the arm cannot reach far enough to grab the hem. The planner also sometimes gives the arms incompatible assignments.

\textbf{Silverware sort} ($60\%$, $12/20$). Failures come from localization errors with multiple objects: robots do not always pursue the objects in their own half of the workspace, causing arm clashes or targets out of reach. Adding another external camera would help here.

\subsection*{UR10e: live displacement and timing}

After the plan is written, the operator moves the object to pick, the destination, or both by hand, by up to about $70$\,cm. The robot detects each object again before the step that names it and keeps the plan; $36$ of the $40$ trials ran on a single plan. The planner is the Qwen3.8-27B planner of Table~\ref{tab:libero_main} on a separate GPU, and the operator judges success. In both failed Utensils in bowl trials the depth of the moved spoon read about $7$ to $9$\,cm too high on the reflective metal, so the gripper closed above it.

Table~\ref{tab:physical_live} gives successes and timing for the static and the live-displacement trials. Trial time runs from the start of execution to the end of the final check, so it includes planning, and it grows with the number of pick-and-place cycles (about $18$\,s per piece in Utensils in tray). The planner is called once per trial unless an attempt fails, detection runs before each step that names an object, and the control loop runs at $500$\,Hz.

\begin{table}[h]
\centering
\caption{UR10e trials with the Qwen3.8-27B planner: static ($20$ per task) and live displacement ($10$ per task). Times and shares are medians and counts are means per trial; memory is allocated (reserved).}
\label{tab:physical_live}
\small
\setlength{\tabcolsep}{3pt}
\begin{tabular}{lcccc}
\toprule
 & Utensils in tray & Stack blocks & Plushie in bowl & Utensils in bowl \\
\midrule
\multicolumn{5}{l}{\emph{Static}} \\
Successes & $20/20$ & $19/20$ & $20/20$ & $18/20$ \\
Trial time & $47.8$\,s & $30.1$\,s & $31.1$\,s & $38.8$\,s \\
Plans per trial & $1.00$ & $1.05$ & $1.10$ & $1.15$ \\
Detection calls per trial & $10.0$ & $7.3$ & $6.2$ & $7.2$ \\
Detection calls after a failed check & $0.15$ & $0.10$ & $0.40$ & $0.65$ \\
Time per detection call & $0.51$\,s & $0.48$\,s & $0.52$\,s & $0.58$\,s \\
Share of trial time in perception & $10.6\%$ & $8.9\%$ & $6.6\%$ & $6.7\%$ \\
Peak GPU memory, perception (GB) & $4.8$ ($8.5$) & $5.0$ ($8.8$) & $4.7$ ($5.7$) & $4.6$ ($8.2$) \\
\midrule
\multicolumn{5}{l}{\emph{Live displacement}} \\
Successes & $10/10$ & $10/10$ & $10/10$ & $8/10$ \\
Trial time & $54.4$\,s & $37.4$\,s & $43.5$\,s & $55.7$\,s \\
Plans per trial & $1.00$ & $1.00$ & $1.20$ & $1.30$ \\
Detection calls per trial & $17.6$ & $8.4$ & $9.7$ & $9.3$ \\
Detection calls after a failed check & $0.10$ & $0$ & $0.70$ & $0.70$ \\
Time per detection call & $0.52$\,s & $0.48$\,s & $0.50$\,s & $0.52$\,s \\
Share of trial time in perception & $15.6\%$ & $10.1\%$ & $10.0\%$ & $5.4\%$ \\
Peak GPU memory, perception (GB) & $4.8$ ($8.6$) & $4.9$ ($8.0$) & $4.7$ ($5.5$) & $4.6$ ($8.0$) \\
Moves found (median size) & $27$ ($20$\,cm) & $11$ ($33$\,cm) & $14$ ($21$\,cm) & $14$ ($18$\,cm) \\
\bottomrule
\end{tabular}
\end{table}

\paragraph{Grasp strategy on thin objects.}
The default top-down grasp strategy sets the gripper yaw to be aligned with the oriented bounding box (principal axis) of the object mask. This results in a clean bounding box when applied to a flat tray. Applied to a thin knife or spoon, however, it resolves to a skinny sliver whose principal axis is unstable and flips between frames (Figure~\ref{fig:failures}f), causing early utensil grasps to either approach across the blade or miss the utensil. Silverware plans can therefore replace the mask-derived yaw with \texttt{grasp\_se3}, which selects a full SE(3) grasp pose from an SE(3)-equivariant grasp generator~\citep{lim2024equigraspflow} and then reaches it using PyRoki's IK~\citep{kim2025pyroki}. In the UR10e trials of Table~\ref{tab:physical} the plans keep the bounding-box yaw. When the depth points of a utensil are unreliable, the executor takes the yaw from the outline of the mask.

\section{Example Behavior Trees}
\label{sec:appendix:bt-examples}

These are generated by Gemini based on SAM3 detection of real world environments. The parameters are what Gemini outputs (we condensed the YAML for readability). Sorting silverware is an example of Gemini's plan when told to operate in bimanual mode.

\noindent
\begin{minipage}[t]{0.47\linewidth}
\subsection*{Single-Arm: Scrub a Plate (FR3)}

With the ability to use primitive actions and our \texttt{constrained\_scrub} skill Gemini is able to grab the sponge and scrub the plate in a circular motion. It will return the sponge to its initial position.

{\fontsize{6.7}{8.2}\selectfont
\begin{verbatim}
task: Pick up the sponge, scrub the plate in a
      circular pattern, and return the sponge
tree:
  type: sequence
  children:
  - type: grasp_se3
    params:
      keypoint_label: sponge
      strategy: top_down
      force: 25
      target_width: 0.035
  - type: move_relative
    params:
      dx: 0.0
      dy: 0.0
      dz: 0.1
  - type: constrained_scrub
    params:
      target_label: plate
      force_n: 8.0
      n_cycles: 4
      scrub_radius_m: 0.04
      pattern: circle
  - type: move_relative
    params:
      dx: 0.0
      dy: 0.0
      dz: 0.1
  - type: move_to_keypoint
    params:
      keypoint_label: sponge
      offset_x: 0.0
      offset_y: 0.0
      offset_z: 0.02
  - type: release
\end{verbatim}}
\end{minipage}\hfill
\begin{minipage}[t]{0.51\linewidth}
\subsection*{Bimanual: Sort Silverware (Panda + FR3)}

Using the \texttt{parallel} node Gemini is able to create two subtrees with the \texttt{arm} tag on every primitive action, which join at an implicit barrier. The left arm grabs the knife while the right arm grabs the fork, and then they both place them in the tray.

{\fontsize{6.7}{8.2}\selectfont
\begin{verbatim}
task: place the fork and the knife into the
      tray, using both arms
tree:
  type: sequence
  children:
  - type: parallel
    children:
    - type: sequence
      children:
      - type: pick_with_arm
        params: {arm: left, keypoint_label: knife,
                 target_width: 0.01, force: 40,
                 prefer_angled: false}
      - type: move_relative
        params: {dx: 0.0, dy: 0.0, dz: 0.2}
    - type: sequence
      children:
      - type: pick_with_arm
        params: {arm: right, keypoint_label: fork,
                 target_width: 0.01, force: 40,
                 prefer_angled: false}
      - type: move_relative
        params: {dx: 0.0, dy: 0.0, dz: 0.2}
  - type: parallel
    children:
    - type: sequence
      children:
      - type: move_to_keypoint_arm
        params: {arm: left, keypoint_label: tray,
                 offset_x: 0.0, offset_y: 0.05,
                 offset_z: 0.08}
      - type: place_with_arm
        params: {arm: left, tilt_angle: 0.0}
    - type: sequence
      children:
      - type: move_to_keypoint_arm
        params: {arm: right, keypoint_label: tray,
                 offset_x: 0.0, offset_y: -0.05,
                 offset_z: 0.08}
      - type: place_with_arm
        params: {arm: right, tilt_angle: 0.0}
\end{verbatim}}
\end{minipage}

\section{Details and Per-Task Results for Table~\ref{tab:libero_main}}
\label{sec:appendix:provenance}

The planner is not given BDDL, and no stored programs are retrieved or saved. Perception uses alternative prompts and the grounding rules described in \S\ref{sec:adaptive} for verifying the wrist and relations. No ground-truth object poses are read from the simulator during rollout. The planner can write grasp and placement checks into the plan, and they act through its retry and fallback nodes (\S\ref{sec:fallback}). After a failed attempt the robot retracts, detects again and asks for a new plan, at most 2 times. The two rows differ only in what decides that an attempt failed (Appendix~\S\ref{sec:appendix:exec}). We use Qwen3.8-27B quantized to 4-bit with AutoRound~\citep{cheng2024autoround} and served with vLLM~\citep{kwon2023vllm} in every trial (Appendix~\S\ref{sec:appendix:gemini}). Tasks and initial states from \libpro are used directly. Trial $i$ loads initial state $i$, and we do not reset the environment after loading it. All cells have $10$ tasks $\times$ $50$ trials, so each row pools $3{,}000$ trials. The 95\% intervals in Table~\ref{tab:libero_main} resample tasks within each setting. In a paired comparison on the same tasks the \rats protocol is $3.4$ points higher $[0.9, 6.4]$. We seed the detect prompt using the objects listed in the perturbed task files and a static set of phrases per object. Scoring uses the success check described in Appendix~\S\ref{sec:appendix:exec}, read at the end of the episode.

The phrases and a $5$\,cm limit on how far the wrist camera can move a pick target were picked on re-sampled placements of the position and task settings: the runner loads a scored initial state and resets the environment, which moves objects by $0.8$--$3.4$\,cm. On these re-sampled placements, taking these two out gives $67.3\%$, compared to $74.0\%$ with them (same $3{,}000$ trials). The language and object perturbations were not used when developing the grounding rules. On these, under the \rats protocol on the $50$ official initial states, we get $69.7\%$ (object $92.4$ and $77.4$, goal $64.6$ and $47.2$, spatial $50.8$ and $85.8$). Our MolmoAct2 numbers use the re-sampled placements.

We measure the time per episode with one simulation worker on the setup of Table~\ref{tab:libero_main}. The planner runs on an RTX 5090, and the worker (simulation, SAM3 and executor) runs on an RTX 4090. Each protocol has $30$ trials ($5$ initial states of one task in each of the six settings). Per trial, the strict protocol has a median wall time of $17.9$\,s from the beginning of the episode to reading the success check, a median planner time of $9.9$\,s, a median perception time of $1.7$\,s and a mean of $2.67$ language-model calls. Under the \rats protocol the medians are $18.8$\,s, $10.1$\,s and $1.7$\,s, with a mean of $3.17$ calls. One planner request takes $2.9$\,s.

In the goal suite, our position losses are concentrated in the middle-drawer and wine-rack tasks, which succeed in $11$ and $3$ out of $50$ trials under the \rats protocol and $7$ and $0$ under the strict protocol. The plate push succeeds in $34$ position trials under the \rats protocol and in $12$ under the strict protocol. Task losses are largest on the rows for the cream cheese in the bowl, the wine bottle on top of the cabinet and the plate push ($11$, $11$ and $15$ successes out of $50$ under the \rats protocol; $12$, $11$ and $2$ strict, and the wine rack falls from $35$ to $2$ strict; Table~\ref{tab:appendix:goal_v5}).

\begin{table}[h]
\centering\scriptsize
\caption{Per-task successes of 50 trials for Table~\ref{tab:libero_main}, object suite (pick up the object and place it in the basket). R: \rats protocol, S: strict. Lang and Obj are the held-out language and object perturbations, under the \rats protocol.}
\label{tab:appendix:object_v5}
\begin{tabular}{lcccccc}
\toprule
 & \multicolumn{2}{c}{Pos} & \multicolumn{2}{c}{Task} & & \\
\cmidrule(lr){2-3}\cmidrule(lr){4-5}
Task & R & S & R & S & Lang & Obj \\
\midrule
Alphabet soup & 41 & 44 & 49 & 50 & 45 & 2 \\
Cream cheese & 50 & 50 & 50 & 49 & 50 & 44 \\
Salad dressing & 50 & 50 & 45 & 46 & 42 & 47 \\
BBQ sauce & 50 & 50 & 50 & 50 & 42 & 45 \\
Ketchup & 13 & 0 & 50 & 50 & 38 & 17 \\
Tomato sauce & 37 & 33 & 50 & 47 & 45 & 35 \\
Butter & 50 & 50 & 50 & 50 & 50 & 50 \\
Milk & 48 & 45 & 37 & 34 & 50 & 47 \\
Chocolate pudding & 50 & 48 & 50 & 48 & 50 & 50 \\
Orange juice & 50 & 50 & 49 & 49 & 50 & 50 \\
\bottomrule
\end{tabular}
\end{table}

\begin{table}[h]
\centering\scriptsize
\caption{As Table~\ref{tab:appendix:object_v5}, spatial suite (pick up the black bowl at the given place and put it on the plate).}
\label{tab:appendix:spatial_v5}
\begin{tabular}{lcccccc}
\toprule
 & \multicolumn{2}{c}{Pos} & \multicolumn{2}{c}{Task} & & \\
\cmidrule(lr){2-3}\cmidrule(lr){4-5}
Task & R & S & R & S & Lang & Obj \\
\midrule
Between the plate and the ramekin & 49 & 49 & 45 & 44 & 3 & 47 \\
Next to the ramekin & 31 & 35 & 49 & 49 & 50 & 31 \\
Table center & 49 & 50 & 45 & 45 & 23 & 47 \\
On the cookie box & 34 & 36 & 45 & 45 & 34 & 50 \\
In the top drawer of the cabinet & 14 & 13 & 29 & 27 & 16 & 11 \\
On the ramekin & 48 & 48 & 13 & 11 & 44 & 47 \\
Next to the cookie box & 20 & 21 & 50 & 50 & 2 & 50 \\
On the stove & 50 & 50 & 49 & 47 & 16 & 49 \\
Next to the plate & 49 & 48 & 46 & 46 & 41 & 48 \\
On the wooden cabinet & 46 & 44 & 50 & 50 & 25 & 49 \\
\bottomrule
\end{tabular}
\end{table}

\begin{table}[h]
\centering\scriptsize
\caption{As Table~\ref{tab:appendix:object_v5}, goal suite.}
\label{tab:appendix:goal_v5}
\begin{tabular}{lcccccc}
\toprule
 & \multicolumn{2}{c}{Pos} & \multicolumn{2}{c}{Task} & & \\
\cmidrule(lr){2-3}\cmidrule(lr){4-5}
Task & R & S & R & S & Lang & Obj \\
\midrule
Open the middle drawer & 11 & 7 & 33 & 32 & 49 & 49 \\
Bowl on the stove & 44 & 42 & 39 & 40 & 50 & 5 \\
Wine bottle on top of the cabinet & 45 & 46 & 11 & 11 & 42 & 49 \\
Open the top drawer, bowl inside & 45 & 47 & 45 & 45 & 9 & 16 \\
Bowl on top of the cabinet & 43 & 40 & 21 & 23 & 50 & 5 \\
Push the plate to the front of the stove & 34 & 12 & 15 & 2 & 0 & 43 \\
Cream cheese in the bowl & 33 & 32 & 11 & 12 & 27 & 18 \\
Turn on the stove & 35 & 35 & 17 & 17 & 41 & 49 \\
Bowl on the plate & 50 & 50 & 34 & 37 & 49 & 0 \\
Wine bottle on the rack & 3 & 0 & 35 & 2 & 6 & 2 \\
\bottomrule
\end{tabular}
\end{table}

\section{Perception-Sourcing Ablation (Ablation Configuration)}
\label{sec:appendix:ablation}
Per-task vectors behind these suite means are in Appendix~\S\ref{sec:appendix:per-task}. All rows share the same planner and controller so they can be compared. \emph{Fair}: task language only, matching \capx. \emph{+BDDL}: adds canonical object names from the tasks' Behavior Domain Definition Language (BDDL) files. \emph{Adaptive}: alternative prompts with SAM3 self-selection.
\begin{table}[h]\centering\small
\begin{tabular}{lccccccc}
\toprule
& \multicolumn{2}{c}{Object} & \multicolumn{2}{c}{Goal} & \multicolumn{2}{c}{Spatial} & \\
Config & Pos & Task & Pos & Task & Pos & Task & Mean \\
\midrule
Fair & $36.4$ & $23.4$ & $36.4$ & $22.4$ & $30.0$ & $43.0$ & $31.9$ \\
+BDDL & $35.0$ & $24.0$ & $36.2$ & $19.4$ & $29.2$ & $43.4$ & $31.2$ \\
Adaptive & $39.8$ & $36.4$ & $40.0$ & $14.0$ & $53.6$ & $72.4$ & $42.7$ \\
\bottomrule
\end{tabular}
\end{table}

\section{Perception Fault Sweep}
\label{sec:appendix:faults}

This sweep measures how success changes when one perception input has a fault. We run all six position and task settings under the \rats protocol on re-sampled placements ($10$ tasks $\times$ $5$ trials per setting, $300$ trials per arm). The clean arm succeeds in $76.7\%$ of trials, and every other arm adds one fault to it. Depth noise is perturbed with $\sigma = kz$, where $z$ is the true depth on every frame; masks are eroded or dilated; or we offset the camera pose used for back-projection with $0.5^\circ$ of rotation error added per cm of offset. Table~\ref{tab:faults} shows the paired difference from the clean arm, with tasks resampled within each setting.

\begin{table}[h]
\centering
\caption{Perception fault sweep on \libpro, \rats protocol, re-sampled placements ($300$ trials per arm). Difference from the clean arm ($76.7\%$) in percentage points, with a 95\% task-bootstrap interval.}
\label{tab:faults}
\small
\begin{tabular}{llcc}
\toprule
Fault & Level & Success ($\%$) & vs.\ clean \\
\midrule
Depth noise, $\sigma = kz$ & $k = 0.01$ & $73.3$ & $-3.3$ $[-8.3, +1.0]$ \\
 & $k = 0.02$ & $67.3$ & $-9.3$ $[-17.3, -2.0]$ \\
 & $k = 0.05$ & $29.0$ & $-47.7$ $[-58.0, -36.3]$ \\
Mask erosion or dilation & $2$\,px & $68.7$ & $-8.0$ $[-12.3, -3.7]$ \\
 & $4$\,px & $64.3$ & $-12.3$ $[-18.7, -6.0]$ \\
 & $8$\,px & $52.0$ & $-24.7$ $[-32.0, -17.3]$ \\
Extrinsic offset ($+0.5^\circ$ per cm) & $0.5$\,cm & $73.0$ & $-3.7$ $[-7.3, 0.0]$ \\
 & $1$\,cm & $70.7$ & $-6.0$ $[-11.0, -1.3]$ \\
 & $2$\,cm & $47.0$ & $-29.7$ $[-37.0, -22.3]$ \\
Scene camera only (wrist camera off) & & $73.7$ & $-3.0$ $[-9.0, +2.7]$ \\
\bottomrule
\end{tabular}
\end{table}

For the mild depth and extrinsic faults and for the scene camera alone, the interval includes $0$, while a mask error of $2$\,px already lowers success. Larger errors degrade performance substantially, by $25$ to $48$ points, since every target that guides a skill is obtained from back-projection. The robot needs calibration within about $1$\,cm, and can operate using only the scene camera.

\section{LIBERO-Displace: Displacement Inside the Episode}
\label{sec:libero_dyn}

All perturbations applied in \libpro and LIBERO-Plus occur prior to the start of an episode, so the scene is static throughout. A policy that memorized a trajectory would still be acting on a static scene. Here we define LIBERO-Displace, in which objects are displaced during execution of the task. THE COLOSSEUM~\citep{pumacay2024colosseum} is a precedent for this format: it extends RLBench, keeps $20$ of its tasks and adds $14$ perturbation factors. We slide the target object along the table at one of $3$ times (after planning before first movement, halfway through the initial approach, just before closing the gripper). We distribute the motion over $40$ sim steps to ensure that the object is moving when the robot is executing. We test magnitudes of $0$, $2$, $5$, $10$\,cm and have a fixed list of eight compass directions the object can travel. The direction is chosen by seeding off the task id and trial number, ensuring all methods see the same sequence of perturbations. The scripted plans of Table~\ref{tab:dyn_main} come from a fixed rule in code with no language model: move to the object, grasp, lift $20$\,cm, move to the destination and release. In every arm of Table~\ref{tab:dyn_main} a detection that is older than $0.5$\,s ($50$\,ms in the arms with scene diff) is replaced before a keypoint move, and the ladder makes at most $3$ attempts. In Table~\ref{tab:libero_dyn} we also show the instant version where the object teleports to its final location in $1$ step. Sliding is the more difficult and more physical scenario. We measure success, number of recoveries and the amount of time it took to detect (perturb to detecting a perturbation) and adapt (detecting a perturbation to resuming normal behavior). The $0$\,cm column is important to ensure your arm can solve the task if nothing changes, since an arm can score well on the perturbed cells and still fail on the undisturbed scene (Table~\ref{tab:libero_dyn}).

If the success check after each attempt is also allowed to trigger recovery, under the \rats protocol, the scripted plans using scene diff and the ladder manage $111$ out of $150$ successes at $10$\,cm displacement over the same $10$ tasks. With neither of these features, they succeed in only $2$ out of $150$.

\paragraph{Ladder against replan.} Table~\ref{tab:dyn_matched} compares the ladder to replan on the same displaced trials, both following plans produced by the planner; there are $150$ trials in each cell, all under the strict protocol, in a separate run of these arms from Table~\ref{tab:dyn_main}. The advantage of the ladder over replan is $10.0$ points at $5$\,cm displacement ($[2.0, 20.0]$) and $12.7$ at $10$\,cm ($[0.0, 32.0]$), requiring on average $1.14$ versus $1.56$ calls to the planner per displaced trial. There is little difference between them with no displacement, over $600$ paired trials covering all six \libpro settings: $457$ versus $464$ under the \rats protocol ($-1.2$ points, $[-3.3, 0.8]$) and $441$ versus $443$ under the strict protocol, and $426$ with recovery turned off. On average the ladder makes $2.71$ calls to the language model per trial, and replan $3.15$.

\begin{table}[t]
\centering
\caption{Ladder against replan under object displacement, planner-written plans, strict protocol. Each entry is successful trials out of $150$ at $0$ / $2$ / $5$ / $10$\,cm (10 object-suite tasks, 3 displacement phases, 5 initial states).}
\label{tab:dyn_matched}
\small
\begin{tabular}{lcc}
\toprule
Recovery & No scene diff & With scene diff \\
\midrule
Replan & $120$ / $104$ / $70$ / $72$ & $123$ / $108$ / $84$ / $89$ \\
Ladder & $120$ / $111$ / $85$ / $91$ & $123$ / $114$ / $95$ / $105$ \\
\bottomrule
\end{tabular}
\end{table}

\begin{table}[t]
\centering
\caption{LIBERO-Displace successes of $45$ trials per cell (three object-suite tasks $\times$ three injection phases $\times$ five initial states), with identical displacement schedules within each block. The two \spark arms execute the same scripted plans. The VLA rows are LIBERO-finetuned checkpoints under the instantaneous injector; their $0$\,cm controls (OpenVLA $86.7\%$, $\pi_{0.5}$ $93.3\%$) match their published object-suite averages. Recov.\ counts logged recovery events over the perturbed cells.}
\label{tab:libero_dyn}
\setlength{\tabcolsep}{4pt}
\begin{tabular}{lccccc}
\toprule
Arm & $0$\,cm & $2$\,cm & $5$\,cm & $10$\,cm & Recov. \\
\midrule
\multicolumn{6}{l}{\emph{Continuous slide (40 steps)}} \\
\spark, verify and adapt & $39/45$ & $37/45$ & $35/45$ & $30/45$ & $185$ \\
\spark, no verification  & $33/45$ & $26/45$ & $7/45$  & $1/45$  & $0$ \\
\midrule
\multicolumn{6}{l}{\emph{Instantaneous displacement (one state write)}} \\
\spark, verify and adapt & $36/45$ & $19/45$ & $23/45$ & $12/45$ & $135$ \\
\spark, no verification  & $27/45$ & $21/45$ & $1/45$  & $0/45$  & $0$ \\
OpenVLA (finetuned)      & $39/45$ & $25/45$ & $4/45$  & $0/45$  & $0$ \\
$\pi_{0.5}$ (finetuned)  & $42/45$ & $38/45$ & $17/45$ & $5/45$  & $0$ \\
\bottomrule
\end{tabular}
\end{table}

For a slide we see the verify and adapt arm drop $7$ trials ($15.6$ points) from $2$\,cm to $10$\,cm ($37/45$, $35/45$, $30/45$) compared to the no verify arm which drops from $26/45$ to $7/45$ to $1/45$. Because our camera scene diff will detect the object leaving its bound position and our in flight retarget can re-aim during that time step, $44$ out of our $102$ successful perturbed trials go without a detection flag. When there is a problem it's typically at the point of our grasp. If our object is still moving at the time we attempt to grasp it will be obscured by our gripper and our pre-close diff will not detect the movement. Then our grasp telemetry will detect we have not grasped our object and will take longer to detect than a jump would (median of $4.6$\,s vs $2.4$\,s). This is the recovery path behind most of our $185$ logged events. As in the \rats-protocol row of Table~\ref{tab:libero_main}, both blocks read the success check after each attempt, and a failed check can start a recovery. With our instant block we can see similar results just with a steeper drop off. Verify and adapt can keep a success rate of $23/45$ and $12/45$ at $5$\,cm and $10$\,cm respectively. Our score at $2$\,cm is only $19/45$ which is lower than the VLA rows and our no verify arm. This is because a $2$\,cm jump usually falls within our gripper's ability to grab an object, so a triggered re-detection costs more than it gains. $\pi_{0.5}$ was able to get a higher score at $2$\,cm ($38/45$, highest score for any arm). It's able to compensate for smaller movements due to its chunking. Both VLAs drop once the movement is greater than what the gripper can grab ($17/45$ and $4/45$ at $5$\,cm, $5/45$ and $0/45$ at $10$\,cm) and they will not know that they failed. 

There is no detect, adapt or recover time for either policy, since neither has a failure flag or a way to replan. $\pi_{0.5}$ also fails on the $5$\,cm and $10$\,cm movement after planning ($5/15$ and $1/15$), even though this movement occurs before the model has seen the scene. The movement places the object outside of the placements it was trained on, so distribution shift causes failure on top of any surprise mid-episode. Detect using our instant injection was $0.34$\,s ($n{=}66$) using our camera scene diff and $2.39$\,s ($n{=}64$) using our grasp telemetry. Adapt took $0.000$\,s ($n{=}50$) to use our in flight retarget and resume our trajectory, and $2.29$--$2.40$\,s ($n{=}77$) to use our recovery subtree and retract our arm before detecting. Detecting using our scene diff was $0.53$\,s ($n{=}40$) and $4.62$\,s ($n{=}51$) using telemetry. Adapting using our retarget took $0.000$\,s ($n{=}41$) and $3.52$--$3.85$\,s ($n{=}45$) to use our recovery. Two instances out of $180$ of our adaptive trials had our post condition met on a failed trial. The object was less than $10$\,cm from our place location in $x$ and $y$ but did not make it into the container, which an $xy$ predicate can't see.

\end{document}